\pdfoutput=1

\documentclass[11pt]{article}

\usepackage{acl}
\usepackage{comment}
\usepackage{times}
\usepackage{latexsym}
\usepackage{graphicx}
\usepackage{amsmath, amssymb} 
\usepackage{comment}
\usepackage{booktabs}
\usepackage{subcaption}
\usepackage{xspace}

\usepackage[T1]{fontenc}

\usepackage[utf8]{inputenc}

\usepackage{microtype}

\usepackage{inconsolata}
\usepackage[textsize=tiny]{todonotes}
\usepackage{footmisc}
\newcommand\blfootnote[1]{%
  \begingroup
  \renewcommand\thefootnote{}%
  \footnotetext{#1}%
  \endgroup
}
\title{Reverse-Engineering the Reader}

\usepackage{rycolab}
\DeclareMathOperator*{\E}{\mathbb{E}}

\newcommand{\mymacro}[1]{{#1}}

\newcommand{\eos}{{\mymacro{\textsc{eos}}}}
\newcommand{\kl}{\mymacro{\mathrm{KL}}}
\newcommand{\reward}{\mymacro{\mathrm{r}}}
\newcommand{\penaltyterm}{\mymacro{\varphi}}

\newcommand{\unit}{\mymacro{u}}
\newcommand{\units}{\mymacro{\boldsymbol{u}}}

\newcommand{\ridgeReg}{\mymacro{\gamma}}

\newcommand{\choleskyL}{\mymacro{\mathrm{\mathbf{L}}}}

\newcommand{\choleskyz}{\mymacro{\mathrm{\mathbf{z}}}}

\newcommand{\klReg}{\mymacro{\lambda}}

\newcommand{\context}{\mymacro{\boldsymbol{c}}}
\newcommand{\surprisal}{\mymacro{\iota}}

\newcommand{\predvec}{\mymacro{\mathbf{x}}}

\newcommand{\lossPSY}{\mymacro{\mathcal{J}}}

\newcommand{\dataset}{\mymacro{\mathcal{D}}}

\newcommand{\btheta}{\boldsymbol{\mymacro{\theta}}}
\newcommand{\bbeta}{\mymacro{\boldsymbol{\mathrm{\beta}}}}

\newcommand{\psychometric}{\mymacro{\psi}}
\newcommand{\bpsychometric}{\mathrm{\boldsymbol{\psychometric}}}

\newcommand{\predpsychometricbetaRegTrg}{\mymacro{\widehat{\psi}_{\betaRegTrg}}}

\newcommand{\betaRegBl}{\bbeta_{\text{b}}}
\newcommand{\betaRegTrg}{\bbeta_{\btheta}}
\newcommand{\betaModel}{\betaRegTrg^{\star}} 

\newcommand{\lin}{\mymacro{f_{\betaRegTrg}}}
\newcommand{\linBase}{\mymacro{f_{\betaRegBl}}}

\newcommand{\model}{\mymacro{p_{\btheta}}}

\newcommand{\prompt}{\mymacro{\boldsymbol{c}}}

\newcommand{\sampleString}{\mymacro{\boldsymbol{u}}}
\newcommand{\approxReward}{\mymacro{\widetilde{\reward}}}

\newcommand{\surprisalvariance}{\text{UID}_{v}}
\newcommand{\localsurprisalvariance}{\text{UID}_{lv}}

\newcommand{\id}{\mymacro{\mathbf{I}}}
\newcommand{\idscale}{\mymacro{\rho}}

\newcommand{\featmat}{\mymacro{\mathbf{X}}_{\btheta}}

\newcommand{\alphabet}{\Sigma}
\newcommand{\eosalphabet}{\overline{\Sigma}}
\newcommand{\alphabetstar}{\alphabet^*
}

\newcommand{\phuman}{{\mymacro{p_{\mathrm{H}}}}}

\newcommand{\pref}{\mymacro{p_{\text{ref}}}}
\newcommand{\prefix}{\mymacro{\pi_{\text{ref}}}}

\newcommand{\prefixp}{\mymacro{\pi_{\btheta}}}
\newcommand{\prefixnormp}{\mymacro{\text{Z}_\prefixp}}

\newcommand{\dll}{\Delta_{\text{llh}}}

\usepackage{amsmath,amsfonts,bm}

\def\eqref#1{equation~\ref{#1}}

\def\1{\bm{1}}

\DeclareMathAlphabet{\mathsfit}{\encodingdefault}{\sfdefault}{m}{sl}
\SetMathAlphabet{\mathsfit}{bold}{\encodingdefault}{\sfdefault}{bx}{n}

\newcommand{\R}{\mathbb{R}}

\newcommand{\defeq}{\mathrel{\stackrel{\textnormal{\tiny def}}{=}}}

\DeclareMathOperator*{\argmin}{argmin}

\author{
  \textbf{Samuel Kiegeland\textsuperscript{\normalfont*1}} \quad
  \textbf{Ethan Gotlieb Wilcox\textsuperscript{\normalfont*1}} \quad
  \textbf{Afra Amini\textsuperscript{\normalfont1}} \quad
  \\
  \textbf{David Robert Reich\textsuperscript{\normalfont2,\normalfont3}} \quad
  \textbf{Ryan Cotterell\textsuperscript{\normalfont1}}
\\
  \textsuperscript{1}ETH Zürich \quad
  \textsuperscript{2}University of Potsdam \quad
  \textsuperscript{3}University of Zürich
\\
\texttt{\href{mailto:skiegeland@ethz.ch}{skiegeland@ethz.ch}} \quad
\texttt{\{\href{mailto:ethan.wilcox@inf.ethz.ch }{ethan.wilcox}, \href{mailto:afra.amini@inf.ethz.ch }{afra.amini}, \href{mailto:afra.amini@inf.ethz.ch }{ryan.cotterell}\}@inf.ethz.ch} \\ \texttt{\href{mailto:david.reich@@uni-potsdam.de}{reich@cl.uzh.ch}}
}

\usepackage[textsize=tiny]{todonotes}
\makeatletter
\newcommand*\iftodonotes{\if@todonotes@disabled\expandafter\@secondoftwo\else\expandafter\@firstoftwo\fi} 
\makeatother

\begin{document}
\maketitle
\blfootnote{*Equal contribution.}
\begin{abstract}

Numerous previous studies have sought to determine to what extent language models, pretrained on natural language text, can serve as useful models of human cognition.
In this paper, we are interested in the opposite question: whether we can directly optimize a language model to be a useful cognitive model by aligning it to human psychometric data.
To achieve this, we introduce a novel alignment technique in which we fine-tune a language model to implicitly optimize the parameters of a linear regressor that directly predicts humans' reading times of in-context linguistic units, e.g., phonemes, morphemes, or words, using surprisal estimates derived from the language model. 
Using words as a test case, we evaluate our technique across multiple model sizes and datasets and find that it improves language models' psychometric predictive power.
However, we find an inverse relationship between psychometric power and a model's performance on downstream NLP tasks as well as its perplexity on held-out test data.
While this latter trend has been observed before \citep{oh2022surprisal, shain2024logrithmic}, we are the first to induce it by manipulating a model's alignment to psychometric data.
\end{abstract}

\section{Introduction} 

Language comprehension is thought to be predictive and incremental. 
Research on reaction times \citep{fischler-et-al-1979-automatic}, fixation patterns \citep{ehrlich-et-al-1981-context}, and brain activations \citep{Kutas1984BrainPD, DeLong2005ProbabilisticWP} suggests that comprehenders anticipate upcoming linguistic units based on the context in which they occur \citep{kuperberg2016prediction}.\footnote{Our code is available at \url{https://github.com/samuki/reverse-engineering-the-reader}.} 
In addition, a large body of evidence shows that when linguistic units are unexpected, they require more cognitive effort to process \citep[\emph{inter alia}]{miller-1964, ehrlich-et-al-1981-context, BALOTA1985364}. 
E.g., reading times of units, e.g., morphemes, words, and sentences, are taken as a measure of cognitive effort, i.e., the less likely the unit is in context, the longer it takes to read \citep{smith2013-log-reading-time}.\looseness=-1

In this study, we are interested in reverse-engineering the part of the language processing system responsible for predicting abstract linguistic units and testing it by measuring its ability to predict reading times.\footnote{In our experiments, we predict abstract units that correspond to words. However, in order to frame our discussion more generally and to align it with our mathematical framework, we use the term \defn{units} instead.} 
To do so, we need, first, to establish a theoretical link between predictability and reading times. 
For this, we draw on \defn{surprisal theory} \citep{hale-2001-probabilistic, levy2008expectation}, which posits that the cognitive effort to process a unit is proportional to its surprisal---the unit's negative log probability given the preceding context.
Implicitly, surprisal theory assumes that a comprehender maintains a probability distribution over upcoming units, i.e., it assumes a \emph{human} language model. However, this human language model is a theoretical construct, and cannot be observed directly. Thus, most previous work that tests surprisal theory has done so using probability estimates derived from a language model trained on large swathes of human-written text.
Under this paradigm, it has been observed that surprisal estimates derived from language models, fit with regularized maximum-likelihood estimation on large corpora, do yield significant predictors although these findings vary based on the quality \citep{goodkind-bicknell-2018-predictive, wilcox2020predictive, wilcox-etal-tacl23}, size, and training-data set size \citep{oh2022surprisal, shain2024logrithmic} of the model. 
One current line of research therefore seeks to uncover what characteristics of pretrained LMs produce better predictors of human reading times, and what this tells us about the human language processing system.\looseness=-1

Our paper asks a simple question: Instead of assessing the ability of pretrained LMs to serve as psycholinguistic predictors, can we directly estimate (or fine-tune) a language model so that its surprisal estimates become better predictors of processing effort for linguistic units? 
We frame this problem as one of aligning the language model to human data \citep{christiano-2017-deep,schulman-2017-proximal, ouyang-etal-2022-training, ziegler-etal-2019-rlhf, rafailov-2023-direct}. However, in contrast to much previous work, which uses human--model alignment to obtain improvement on natural language processing tasks, e.g., summarizing text \citep{stiennon-2020-learning} or producing non-toxic outputs \citep{toxicityhf}, we seek to align LMs to be better psychometric predictors. 
Particularly, we aim to directly align models to human \emph{reading} data by optimizing the parameters of the statistical models typically used to evaluate the psychometric fit. 
While recent approaches like direct preference optimization \citep[DPO;][]{rafailov-2023-direct} are designed to optimize a model's parameters based on human preferences, they rely on pairwise preference data, which is not applicable to real-valued psychometric data.
Thus, we propose a novel alignment technique that allows us to directly optimize the language model's parameters in such a way that it serves as a better predictor of real-valued psychometric data. 
Specifically, we fine-tune the language model to implicitly optimize the coefficients of a linear regression that predicts the reading time of an individual unit.\looseness=-1 

We test our technique on three English-language reading datasets and find that it increases the statistical fit of a linear regressor in terms of the likelihood it assigns to reading times on a held-out test set.
We also observe a positive relationship between a model's psychological predictive power and its perplexity.
While it has been observed that better LMs are better psychological models of reading up to a point \citep{goodkind-bicknell-2018-predictive, wilcox2020predictive, wilcox-etal-2023-language}, after a certain model size, their fit to human reading times decreases \citep{oh2022surprisal, shain2024logrithmic}. In other words, beyond an inflection point, better LMs are \emph{worse} predictors of human reading times. Through our alignment procedure, we are able to demonstrate the contrapositive, namely that as we causally make our language models' outputs more aligned with reading, they become worse at predicting the next word.\looseness=-1

\section{Psycholinguistics Background}
\label{sec:psycholinguistic-background}
Put concisely, the goal of this paper is to reverse-engineer $\phuman$, a person's internal language model from psychometric data collected through experimentation.
Our reasoning is as follows: if we can align an existing language model $\model$ to more accurately predict such psychometric data, $\model$ will also more closely resemble $\phuman$.\looseness=-1

\subsection{Language Models}
Let $\alphabet$ be an \defn{alphabet}, i.e., a finite, non-empty set, and let $\eosalphabet \defeq \alphabet \cup \{\eos\}$ be the alphabet augmented with a distinguished end-of-string symbol not in $\alphabet$.
A \defn{language model} $\model$ is a probability distribution over $\alphabetstar$, which is the set of all strings over $\alphabet$.
Further, following \citet{opedal2024rolecontextreadingtime}, we define the \textbf{normalized prefix probability}\looseness=-1
\begin{equation}
\label{eqn:prefix-probab}
 \prefixp(\context) \defeq \frac{1}{\prefixnormp} \sum_{\units \in \alphabetstar} \model(\context \units),
\end{equation}
which is a probability distribution over prefixes $\context \in \alphabetstar$, where $\context \units$ is the concatenation of $\context$ and $\units$.
The normalization constant $\prefixnormp = 1 + \sum_{\units \in \Sigma^{*}}\model(\units)\vert\units\vert$ ensures that all probabilities sum to one.
Here $\vert\units\vert$ denotes the length, i.e., the number of units in a string $\units$.
Note that \Cref{eqn:prefix-probab} is only well-defined in an LM with finite expected length.

\subsection{Psychometric Measurements}
Let $\psychometric(\unit, \context) \in \R$ denote a measurement for a unit $\unit \in \eosalphabet$ appearing in context $\context \in \alphabetstar$.
In this paper,  $\psychometric(\unit, \context)$ represents various reading time measurements for a given unit, such as gaze duration, first fixation duration, and total fixation duration, which are standard approximations to the processing effort of a linguistic unit in context \citep[\emph{inter alia}]{miller-1964, Just1980-JUSATO, FRAZIER1982178, rayner-1998}.\looseness=-1

\subsection{Surprisal Theory}\label{sec:surprisal-theory}
Surprisal theory furnishes us with an easy-to-compute predictor of processing effort that is derived from a pretrained language model.
Formally, surprisal theory predicts that the time it takes to process a linguistic unit $\unit \in \eosalphabet$ in context $\context \in \alphabetstar$ is an affine\footnote{Previous work has often described the relationship as linear. However, we use affine here due to the additive constant required to link the two.} function of the unit's contextual surprisal under the human language model $\phuman$, defined as
\begin{equation} \label{eq:surprisal}
    \surprisal_{\mathrm{H}}(\unit \mid \context) \defeq  - \log_2 \phuman(\unit \mid \context).
\end{equation}
Surprisal theory has been supported by numerous empirical studies, which have found that surprisal is predictive of reading times across multiple datasets \citep{smith2013-log-reading-time, wilcox2020predictive, shain2024logrithmic}, types of reading time measurements \citep{pimentel-etal-2023}, and languages \citep{kuribayashi-etal-2021-lower, meister-etal-2021-revisiting, wilcox-etal-tacl23}. Typically, $\phuman$ is approximated using a LM estimated from corpus data, i.e., 
we substitute 
\begin{equation}
    \surprisal_{\btheta}(\unit \mid \context) \defeq -\log_2 \model(\unit \mid \context)
\end{equation}
in for $\surprisal_{\mathrm{H}}$ when predicting processing effort.

\subsection{Linear Modeling}\label{sec:linear-modeling}
We now discuss how empirical support for surprisal theory is typically adduced. 
Following previous work, we assume an affine function links a linguistic unit's contextual surprisal and that unit's reading time\footnote{Although, see \citet{hoover2023sampling} for a different perspective on the shape of the linking function.}  \citep{smith2013-log-reading-time, wilcox-etal-tacl23,shain2024logrithmic} and apply linear regression to predict reading times based on contextual surprisal. 
In mathematical jargon, both the psychometric measurements $\psychometric(\unit, \context) \colon \eosalphabet \times \alphabetstar \rightarrow \R$ and our predictors $\predvec_{\btheta}(\unit, \context) \colon \eosalphabet \times \alphabetstar \rightarrow \R^D$ are real-valued random variables.
In the case of the predictor, given a unit $\unit \in \eosalphabet$ and a context $\context \in \alphabetstar$, we define the predictor as a $D$-dimensional real column vector\looseness=-1
\begin{equation}
\label{eqn:predvec}
\predvec_{\btheta}(\unit, \context) = [\surprisal_{\btheta}(\unit \mid \context), x^2, \ldots, x^{D}]^\top,
\end{equation}
which, as depicted, includes our surprisal estimate $\surprisal_{\btheta}(\unit \mid \context)$; the additional variables $x^2, \ldots, x^D$ are considered to be baseline predictors and are chosen at the modeler's discretion depending on what they seek to test.
Given a parameter (column) vector $\betaRegTrg \in \R^D$, we define the following linear model
\begin{subequations}
\label{eqn:linear-model}
\begin{align}
\psychometric(\unit, \context) &\sim \lin(\cdot \mid \predvec_{\btheta}(\unit, \context)) \\
&= \mathcal{N}(\predpsychometricbetaRegTrg(\unit, \context), \sigma^2),
\end{align}
\end{subequations}
where the linear function 
\begin{equation}
    \label{eqn:predpsychometric}
    \predpsychometricbetaRegTrg(\unit, \context) = \predvec_{\btheta}(\unit, \context)^\top\betaRegTrg
\end{equation}
 constitutes the mean and $\sigma^2$ is the variance.\looseness=-1

We evaluate the predictive power of our model by fitting $\lin$ on a training set and measuring the log-likelihood of the test set; higher log-likelihood indicates greater predictive power of the model.
To assess how much surprisal contributes to the predictive power, we fit two regression models based on two predictors. 
The baseline predictor $\predvec_{\text{b}}(\unit, \context)$, defined identically to \Cref{eqn:predvec}, but with the estimated surprisal zeroed out, typically consists of a unit's unigram surprisal, i.e., its negative log unigram probability\footnote{See \citet{opedal2024rolecontextreadingtime} for a detailed explanation of this predictor variable.} and a unit's length (in characters). 
The target predictor, denoted by $\predvec_{\btheta}(\unit, \context)$, includes the same set of baseline predictors together with the estimated surprisal of the unit $\unit$.\looseness=-1

To quantify the predictive power of contextual surprisal, we compute the delta log-likelihood $\dll$ between the two models, which is the average unit-level difference in log-likelihood assigned by the two predictors to the reading time measurements. 
For a single unit--context pair, we compute:
\begin{equation}
\label{eqn:dll}
\begin{aligned}
    &\dll(\unit, \context) = \log \lin(\psychometric(\unit, \context)\mid \predvec_{\btheta}(\unit, \context)) \\ &\quad\quad - \log \linBase(\psychometric(\unit, \context)\mid \predvec_{\text{b}}(\unit, \context)),
\end{aligned}
\end{equation}
where $\betaRegBl$ and $\betaRegTrg$ are the coefficients for the baseline and target models, respectively, and are estimated separately.
Intuitively, a higher $\dll$ indicates that the estimated surprisals contribute more to the predictive power or psychometric accuracy of the model over reading times, compared to the baseline predictors \citep{frank2011insensitivity}.

Having established a metric, delta log-likelihood, to measure how much contextual surprisal contributes to predicting reading times, we are now ready to answer the question we posed at the beginning: Can we fine-tune a language model such that surprisal estimates derived from it become better predictors of reading times?

\section{Aligning LMs to Psychometric Data}
As discussed in \Cref{sec:linear-modeling}, the psychometric predictive power of a language model is typically evaluated by assessing the predictive power of surprisal estimates $\surprisal_{\btheta}(\unit \mid \context)$ of a linguistic unit $\unit$ in a context $\context$ with respect to human reading times. 
Rather than \emph{evaluating} a language model's psychometric predictive power, in this study, we ask whether we can fine-tune language models to \emph{increase} their psychometric predictive power.\looseness=-1

We treat this as an alignment problem \citep{christiano-2017-deep,schulman-2017-proximal, ouyang-etal-2022-training, ziegler-etal-2019-rlhf, rafailov-2023-direct}.
Let $\pref$ denote a pretrained language model that will serve as a reference.
Further, let $\model$ denote the language model we seek to fine-tune, generally initialized to $\pref$.
Our goal is to align $\model$ such that its surprisal estimates are more directly correlated with psychometric data in comparison to $\pref$. 
By means of implicit differentiation, we derive an objective that allows us to perform such psychometric alignment.\looseness=-1

\subsection{Deriving an Objective}
\paragraph{Reward Function.}
We draw inspiration from direct preference optimization, which implicitly optimizes the parameters of a reward model for predicting the outcomes of pairwise comparisons between items.
However, rather than implicitly fitting a Bradley--Terry model \citep{bradley-terry-1952} to model pairwise preferences given by human annotators, we implicitly fit a linear regressor $\lin$ to model psychometric measurements. 
Following the notation developed in \Cref{sec:linear-modeling}, let $\predvec_{\btheta}(\unit, \context)$ denote the predictor vector, which includes the contextual surprisal derived from the model $\model$. 
To optimize the parameters of $\lin$, we define our reward function as the \emph{negative}\footnote{We use a negative sign to ensure that maximizing the reward corresponds to minimizing the prediction error.} minimum of the expected mean squared error (MSE) between the observed psychometric data $\psychometric(\unit, \context)$ and the predicted values $\predpsychometricbetaRegTrg(\unit, \context) = \predvec_{\btheta}(\unit, \context)^\top \betaRegTrg$ as defined in \Cref{eqn:predpsychometric}.
The reward is then given by
\begin{equation}
\label{eqn:reward}
\reward(\btheta) \!\defeq - \min_{\betaRegTrg \in \R^{D}} \! \E_{\substack{(\unit, \context) \sim \prefix}} \big(\psychometric(\unit, \context) - \predpsychometricbetaRegTrg(\unit, \context)\big)^2,
\end{equation}
\noindent where $\prefix$, as defined in \Cref{eqn:prefix-probab}, is the normalized prefix probability. 
Note that under our linear model specified in \Cref{eqn:linear-model}, maximizing $\reward(\btheta)$ is equivalent to maximizing the $\dll$ in \Cref{eqn:dll}.\footnote{This is equivalent to maximum likelihood estimation. We omit $\linBase$ since it does not depend on $\btheta$.}
\paragraph{Regularization with KL Divergence.}
To prevent the fine-tuned model $\model$ from diverging excessively from the pretrained reference model $\pref$, we regularize our objective with the Kullback--Leibler (KL) divergence, as is typically done in RLHF \citep{schulman-2017-proximal, ziegler-etal-2019-rlhf, stiennon-2020-learning, ouyang-etal-2022-training}. 
More specifically, the regularization term is defined as:
\begin{subequations}
\label{eqn:kl}
\begin{align}
\penaltyterm(\btheta) &\defeq \!\E_{\context \sim \prefix}\! \kl\big(\pref(\cdot \mid \context) \mid\mid \model(\cdot \mid \context)\big) \\
&= \!\E_{\context \sim \prefix}\! \sum_{\unit \in \eosalphabet} \pref(\unit \mid \context) \log \frac{\pref(\unit \mid \context)}{\model(\unit \mid \context)}.
\end{align}
\end{subequations}
where $\prefix$ is the normalized prefix probability of the reference distribution $\pref$.

\paragraph{Putting it All Together.}
We now combine the reward and the KL regularization to define an objective for aligning LMs to psychometric data as 
\begin{equation}
\label{eqn:loss}
\lossPSY(\btheta) \defeq \underbrace{\reward(\btheta)}_{\text{reward}} - \underbrace{\klReg \cdot \penaltyterm(\btheta)}_{\text{KL reg.}},
\end{equation}
where $\klReg \geq 0$ is a hyperparameter, which determines the strength of the KL regularization.
Because optimizing $\reward(\btheta)$ corresponds to optimizing $\dll$, $\lossPSY(\btheta)$ trades off better alignment with human psychometric data against the KL divergence from the pretrained model $\pref$.\looseness=-1

\subsection{Approximation of the Reward Function}
\label{ssec:reward-approximation}
In practice, we use a Monte Carlo estimate of $N$ unit--context pairs $(\unit_n, \context_n)\sim\prefix$ to approximate the expectation in \Cref{eqn:reward}.
Let $\bpsychometric = [\psychometric(\unit_1, \context_1), \ldots, \psychometric(\unit_N, \context_N)]^\top\in\R^N$ denote the real column vector of $N$ reading time observations.
Then we define the approximate reward as
\begin{subequations}
\begin{align}
\approxReward(\btheta) &\defeq \! - \!\!\min_{\betaRegTrg \in \R^{D}}\! \!\frac{1}{N}\sum_{n=1}^N \!\big(\psychometric(\unit_n, \context_n) - \predpsychometricbetaRegTrg(\unit_n, \context_n)\big)^2 \\
&= \!-  \!\!\min_{\betaRegTrg \in \R^{D}}\! \frac{1}{N}||\bpsychometric - \featmat \betaRegTrg ||^2,  \label{eq:final-equation} 
\end{align}
\end{subequations}
where $\featmat$ is an $N \times D$ real matrix, with each row corresponding to a predictor vector, as defined in \Cref{eqn:predvec}.
Leveraging a well-known closed-form solution (see \Cref{asec:regression}), we directly compute \Cref{eq:final-equation}.
To ensure that $\featmat^\top\featmat$ is invertible, we add a small regularization term $\idscale\id$ with $\idscale > 0$, leading to the following coefficients
\begin{equation}
\label{eqn:beta-start-def}
  \!\!   \betaModel  = \big(\underbrace{\featmat^{\top}\featmat  + \idscale\id}_{\text{always invertible}}\big)^{-1} \featmat^{\top} \bpsychometric.
\end{equation}
This results in the simple reward term:
\begin{equation}\label{eqn:rewardApprox}
 \approxReward(\btheta) = -\frac{1}{N}||\bpsychometric - \featmat\betaModel  ||^2.
\end{equation}
Notably, the optimal coefficients are now parameterized by the language model's parameters $\btheta$.\footnote{
To compute the optimal coefficients $\betaModel$ in \Cref{eqn:beta-start-def} efficiently, we use the Cholesky decomposition; see \Cref{asec:cholesky}.
}

\begin{figure*}[ht!]
  \includegraphics[width=\linewidth]{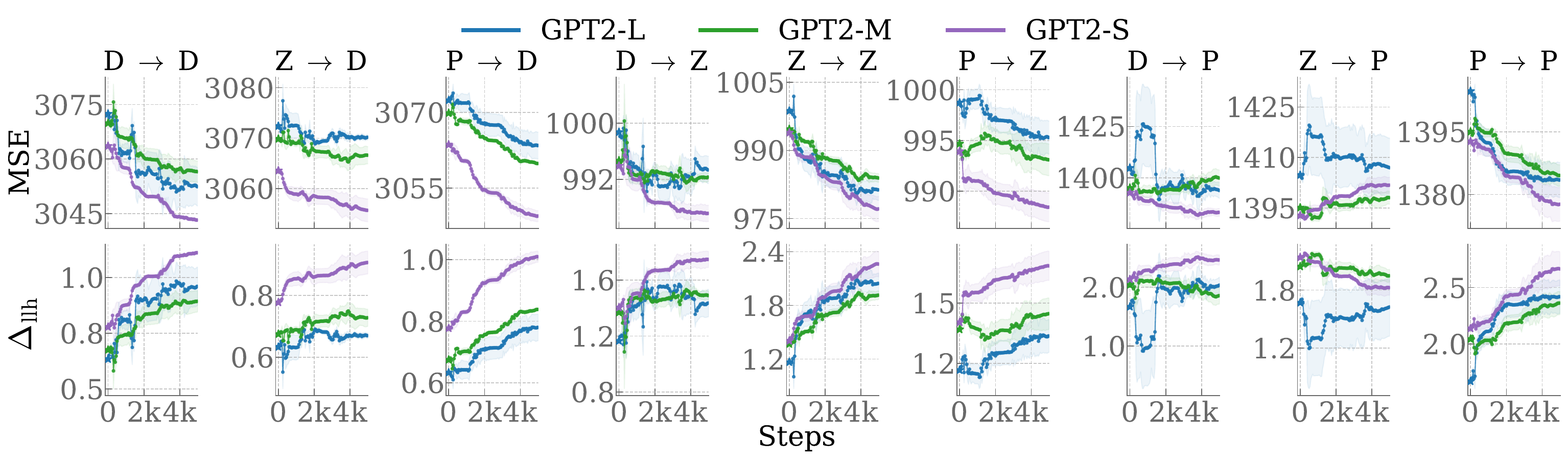}
  \caption{Learning curves for the MSE (top) and $\dll$ ($10^{-2} \text{ nats}$, bottom) on the test datasets throughout fine-tuning.
  Bands show the standard error across random seeds. 
  MSE tends to decrease, while $\dll$ increases, showing better prediction of reading times.
}
  \label{fig:mse-dll-change}
\end{figure*}

\section{Experimental Design} \label{sec:experiments}
In this section, we discuss how we experiment with fine-tuning a language model using the objective defined in \Cref{eqn:loss} and the reward approximation given in \Cref{eqn:rewardApprox}.
Specifically, we design experiments to evaluate the effectiveness of our objective for improving a model's psychometric accuracy, i.e., how well it predicts human reading times. Additionally, we assess the impact of our objective on a model's quality, as measured through its perplexity on test data and its performance on downstream NLP tasks.\looseness=-1

\subsection{Models}
We use the GPT-2 family of models \cite{radford-2019-language} and conduct experiments on the \texttt{small}, \texttt{medium}, and \texttt{large} versions of the model available on the HuggingFace hub~\cite{wolf-etal-2020-transformers}.\looseness=-1

\subsection{Data}
While our objective given in \Cref{eqn:loss} assumes unit--context pairs sampled from $\pref$, we lack psychometric data for LM-generated text.we take to be
However, future work should investigate this assumption more thoroughly.
We fine-tune and evaluate models on three widely used eye-tracking corpora: The \textbf{Dundee Corpus} \citep{kennedy-etal-2003-dundee}, which includes eye-movement data from 10 English-speaking participants reading 2368 sentences of newspaper articles from \textit{The Independent}, the \textbf{Provo Corpus} \cite{luke-etal-2018-provo}, which contains eye-tracking data from 84 participants who read 55 paragraphs of texts from various sources, including fiction and non-fiction, and the \textbf{ZuCo Corpora} (ZuCo 1.0 \citep{hollenstein-etal-2018-zuco1} and ZuCo 2.0 \citep{hollenstein-etal-2020-zuco2}), which contain data from 12 and 18 participants, respectively, reading sentences from Wikipedia articles and movie reviews. \looseness=-1

Similar to previous work \citep{wilcox2020predictive, wilcox-etal-tacl23}, we focus on the gaze duration, defined as the total time of a reader's first pass fixations on a unit $\unit$ before they fixate on a different unit; see \Cref{asec:datasets} for details.
In addition, we conduct experiments using the total reading duration and first fixation duration (\Cref{asec:trt-ffd}). We further verify that our results are not due to random effects through additional experiments with random reading times (\Cref{asec:random-reading}). 
We create test sets by sampling 40\% of the data from each corpus. 
Then, to construct various test sets, we randomly sample 70\% of the remaining 60\% of the data according to 3 random seeds.
We can view this procedure as a simple bootstrapping procedure, from which we can approximate error bars \citep{bootstrap}.
We fine-tune and evaluate models on all pairs of eye-tracking corpora, resulting in 9 unique data splits as shown in \Cref{tab:data_splits_configurations_app}.\looseness=-1

\subsection{Fine-Tuning}
We compute the contextual surprisal of each unit in a sentence, excluding units with zero reading times and zero frequencies, during fine-tuning. 
It  is common practice in psycholinguistics literature to 
drop words that were skipped on the first pass during reading and, therefore, have a first pass reading time of zero \citep{smith2013-log-reading-time, oh2023why}.
Across all experiments, the predictor $\featmat$ consists of a unit's contextual surprisal, its unigram surprisal, estimated using \citeposs{robyn_speer_2022_7199437} toolkit, and its length. 
We opt not to include coefficients for spillover. We do so because reading times in eye-tracking studies have been observed to be less susceptible to spillover than other reading modalities, for example, self-paced reading \citep{shain2021continuous}. 
Although, we acknowledge that this is a limitation of our study. 
For all configurations, we fine-tune models for 5k steps and repeat each run using three different random seeds. 
For an overview of all hyperparameters, see \Cref{asec:trainingparameters}.\looseness=-1

\begin{table}\centering
\small
\begin{tabular}{lcc} 
\toprule Configuration & Train Size & Test Size \\ 
\midrule 
Dundee (D) $\rightarrow$ Provo (P) & 20894.7  & 1114  \\ 
Dundee (D) $\rightarrow$ Dundee (D) & 20894.7  & 20207 \\ 
Dundee (D) $\rightarrow$ ZuCo (Z) & 20894.7 & 7715  \\ 
\midrule 
ZuCo (Z) $\rightarrow$ Provo (P) & 7761 & 1114  \\ 
ZuCo (Z) $\rightarrow$ Dundee (D) & 7761 & 20207 \\ 
ZuCo (Z) $\rightarrow$ ZuCo (Z) & 7761 & 7715  \\ 
\midrule 
Provo (P) $\rightarrow$ Provo (P) & 1113.7 & 1114 \\ 
Provo (P) $\rightarrow$ Dundee (D) & 1113.7 & 20207  \\ 
Provo (P) $\rightarrow$ ZuCo (Z) & 1113.7 & 7715 \\
\bottomrule 
\end{tabular} 
\caption{Data splits and configurations for fine-tuning and evaluation. Numbers indicate the mean number of tokens in each split across our random seeds.} \label{tab:data_splits_configurations} 
\vspace{-10pt}
\end{table}

\subsection{Evaluation}
We evaluate the predictive power of the estimated surprisal values for predicting reading times every 50 steps using our test data splits.
During each evaluation phase, we first obtain surprisal estimates on the test data from the aligned language model.
Using these estimates, we perform a 5-fold cross-validation on the test data, where we fit baseline and target linear regressors using ordinary least squares and use them to compute the unit-level mean $\dll$.

\section{Results}
We now return to our main question: Does our objective align language models more closely to human reading times compared to pretrained models?
We first analyze the effect of maximizing the unregularized reward and later, in \Cref{sec:kl-reg}, extend our analysis to the KL-regularized objective.\looseness=-1

\subsection{Predicting Reading Times}
\label{ssec:predicting_rts}
Are fine-tuned models $\model$ better predictors of reading times compared to pretrained models $\pref$? 
To answer this question, we examine both the mean square error (MSE) and the delta log-likelihood $\dll$ computed at each evaluation step on the test data using cross-validation.
While the $\dll$ is our main metric for measuring a model's psychometric accuracy, we use the MSE to compare the magnitude of prediction errors.
The MSE is computed similarly to the unregularized objective, given in \Cref{eqn:loss}, except we calculate the MSE through cross-validation on the entire test set using linear regression to evaluate the predictive power of surprisal estimates.\looseness=-1

As visualized in \Cref{fig:mse-dll-change}, the MSE values are relatively high because our reading times are in milliseconds; an MSE of 4,000 corresponds to a prediction that is off by only about 1/20$^{\text{th}}$ of a second. 
We observe that the MSE decreases for all models across all held-out datasets over the course of fine-tuning.
An exception to this is the MSE for the data splits Dundee $\rightarrow$ Provo and ZuCo $\rightarrow$ Provo, where we do not observe consistent decreases in MSE, potentially due to the small size of the test set. 
Models evaluated on the ZuCo dataset have the lowest MSE, followed by Provo and Dundee.
Further, in \Cref{fig:mse-dll-change}, we visualize the $\dll$ of regressors evaluated on our test dataset over the course of fine-tuning. 
In line with our observed decreases in MSE, we find that $\dll$ increases on most data splits, with the exception of models fine-tuned on Dundee or ZuCo and evaluated on Provo. Smaller models start with higher $\dll$, which is consistent with previous literature \citep{oh2023why}. 
In \Cref{tab:dll-table}, we compare the percentage increase from each model's start $\dll$ to the maximum $\dll$ it achieves over the course of the fine-tuning, averaged over three random seeds.
Interestingly, GPT2-S shows higher percentage increases compared to GPT2-M, suggesting that model size alone does not fully account for these differences.\looseness=-1
\begin{table}
    \small
    \centering
    \begin{tabular}{lrrr} 
    \toprule 
    Model & GPT2-L & GPT2-M & GPT2-S \\ 
     \midrule  
    D $\rightarrow$ D & 55.291 & 32.822 & 42.880 \\ 
    P $\rightarrow$ D & 23.294 & 24.233 & 29.898 \\ 
    Z $\rightarrow$ D & 9.084  & 11.077 & 16.598 \\ 
    \midrule
    D $\rightarrow$ P & 31.356 & 4.928  & 17.780 \\ 
    P $\rightarrow$ P & 45.273 & 15.621 & 24.769 \\ 
    Z $\rightarrow$ P & 2.116  & 6.460  & 1.999  \\ 
    \midrule
    D $\rightarrow$ Z & 36.577 & 11.950 & 24.502 \\ 
    P $\rightarrow$ Z & 15.012 & 5.730  & 20.094 \\ 
    Z $\rightarrow$ Z & 81.390 & 41.801 & 61.196 \\ 
    \bottomrule
    \end{tabular}
    \caption{Percentage increase between the initial and maximum $\dll$ for each model. For exact start and maximum $\dll$ values, see \Cref{tab:dll-table-app}.}
    \label{tab:dll-table}
    \vspace{-10pt}
\end{table} 

\subsection{Coefficient Estimates}
\begin{figure}
  \includegraphics[width=\linewidth]{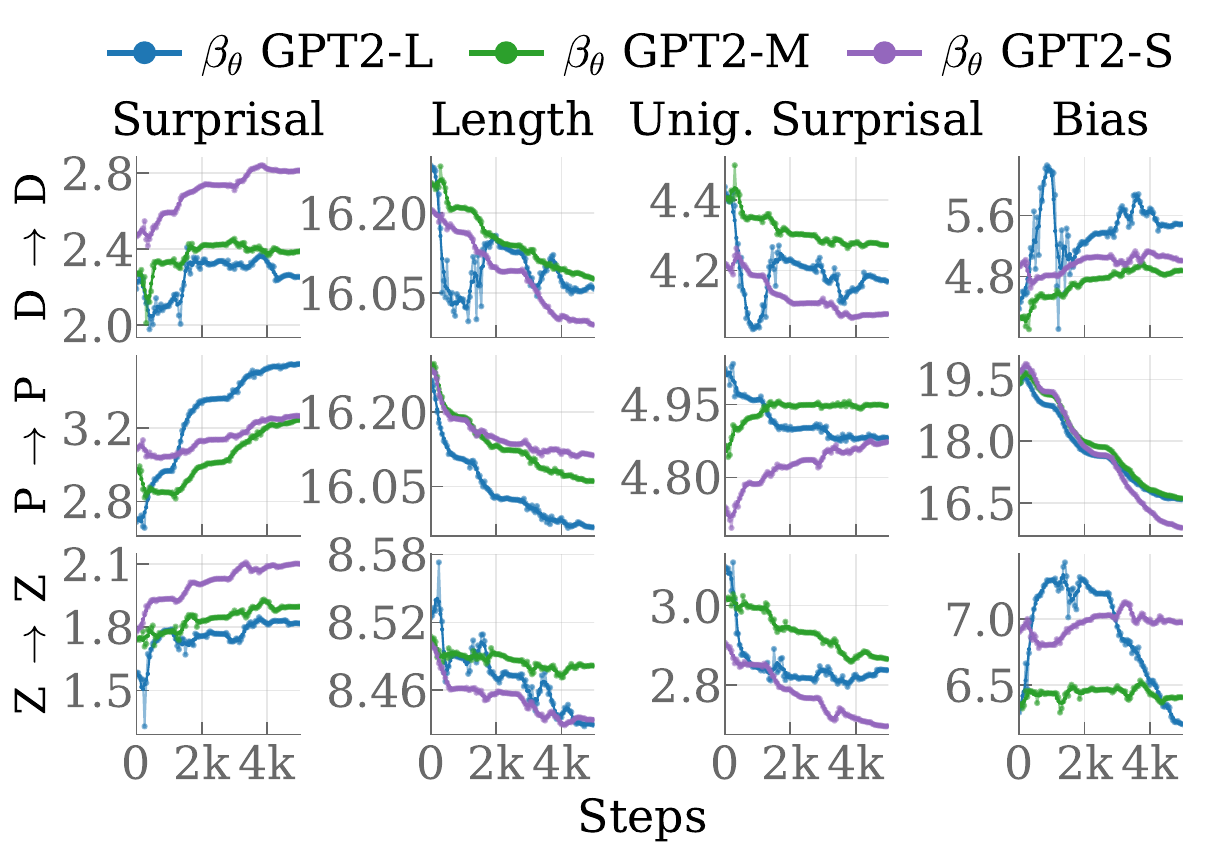}
  \caption {\textbf{Mean Coefficients of unit-level features over fine-tuning.} Smoothed values (window size 5) are shown, with unsmoothed values in a pale version of the color.
  Coefficients corresponding to surprisal tend to increase over the course of fine-tuning.}
  \label{fig:coefficients}
  \vspace{-10pt}
\end{figure}

Additionally, we want to know what the fine-tuned models have implicitly learned about the role of surprisal, as well as the baseline features, over the course of fine-tuning.
To examine this, in \Cref{fig:coefficients}, we visualize the regressor coefficients $\betaRegTrg$ from cross-validation on the whole test set. 
We observe the following tendencies: The coefficients for a unit's contextual surprisal and length are positive, indicating that, as units become less predictable and longer, they take more time to read.

\begin{figure}[ht]
  \includegraphics[width=\linewidth]{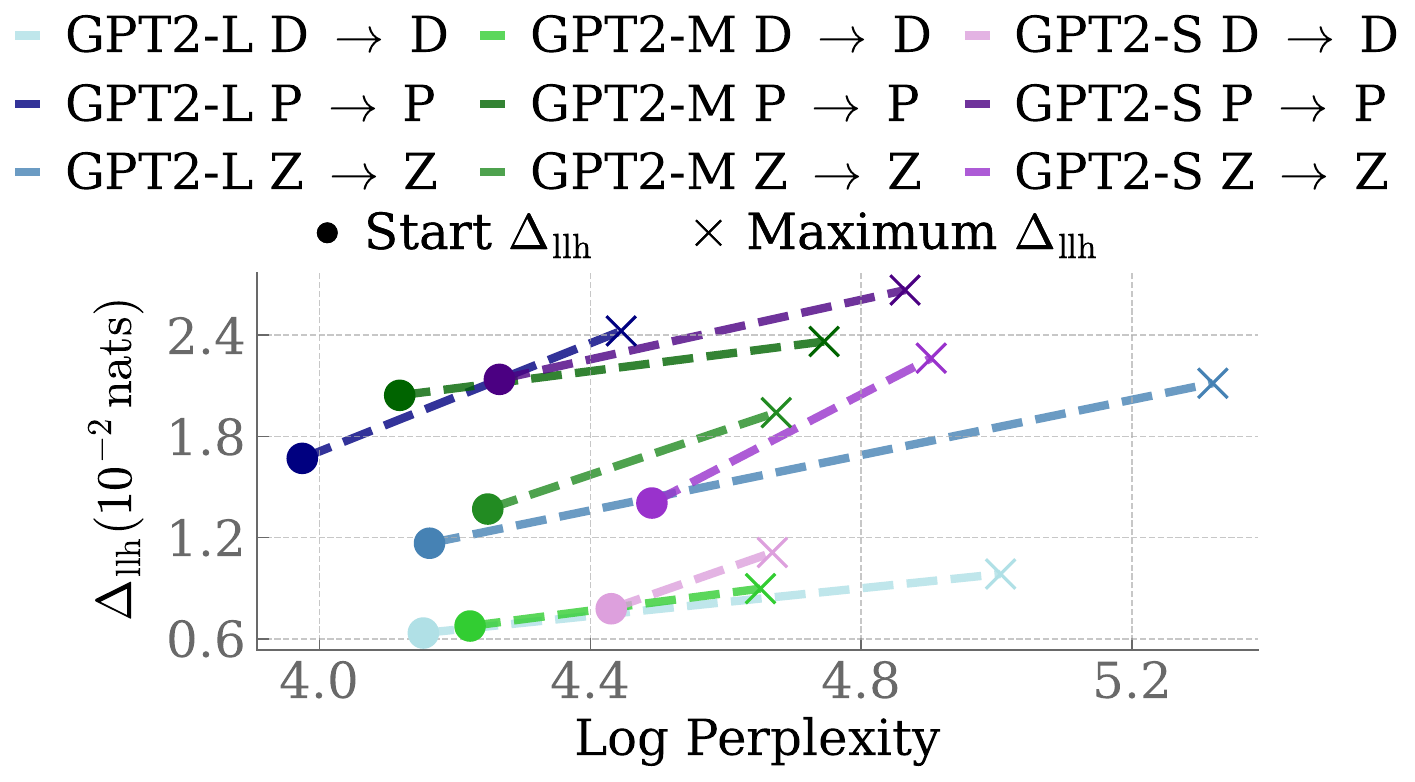}
  \caption{\textbf{Perplexity vs. $\dll$.} We compare model perplexity at the start of fine-tuning to the point where they achieve the highest mean $\dll$. Optimizing models using $\lossPSY$ increases perplexity. See \Cref{fig:dll_ppl_app} for all data splits.\looseness=-1
  }
  \label{fig:perplexity}
  \vspace{-10pt}
\end{figure}

The positive coefficient for unigram surprisal means more frequent units take less time to read. These results align with the previous literature, e.g., those from \citealp{wilcox-etal-tacl23}.
Interestingly, we observe that over the course of fine-tuning, the coefficient for surprisal tends to \emph{increase}, while the coefficient for length \emph{decreases}. 
For unigram surprisal and our bias term, i.e., the intercept of the regression, we observe mixed results, with coefficients for unigram surprisal decreasing for models evaluated on the Dundee and ZuCo datasets and increasing for models evaluated on Provo. 
Overall, the coefficient's trajectories suggest that the predictive power of surprisal for predicting reading times increases throughout fine-tuning.
For coefficients across data splits, as well as coefficients for the regularized objective, see \Cref{asec:all_coeffs}.\looseness=-1

\subsection{Perplexity vs. Quality}
Several recent papers have found that, above a certain size, as a language model's perplexity decreases, its predictive power increases \citep{oh2022surprisal, shain2024logrithmic}. Follow-up work has suggested that this is due to the super-human predictive abilities of the models, especially for low-frequency nouns such as named entities \citep{oh2023why}. 
An open question remains regarding the causal factors behind these dynamics. So far, studies have tested the relationship by manipulating a language model's quality either by choosing different sizes of pretrained models as in \citet{oh2023why} or by training successively smaller and smaller models as in \citet{wilcox-etal-2023-language}. However, our fine-tuning methods allow us to flip the causal arrow. As we make a language model more closely aligned with human reading times, what happens to its quality?
To investigate this question, in \Cref{fig:perplexity}, we show the perplexity and $\dll$ both at the start of fine-tuning and when the model achieves its maximum $\dll$, which is typically near the end of fine-tuning. 
We find that increases in $\dll$ generally correspond to increases in perplexity. These results indicate that as we make a language model's predictions more aligned with reading times, it becomes worse at modeling text.\looseness=-1

\subsection{Kullback--Leibler Regularization}
\label{sec:kl-reg}
\begin{figure*}
  \includegraphics[width=\linewidth]{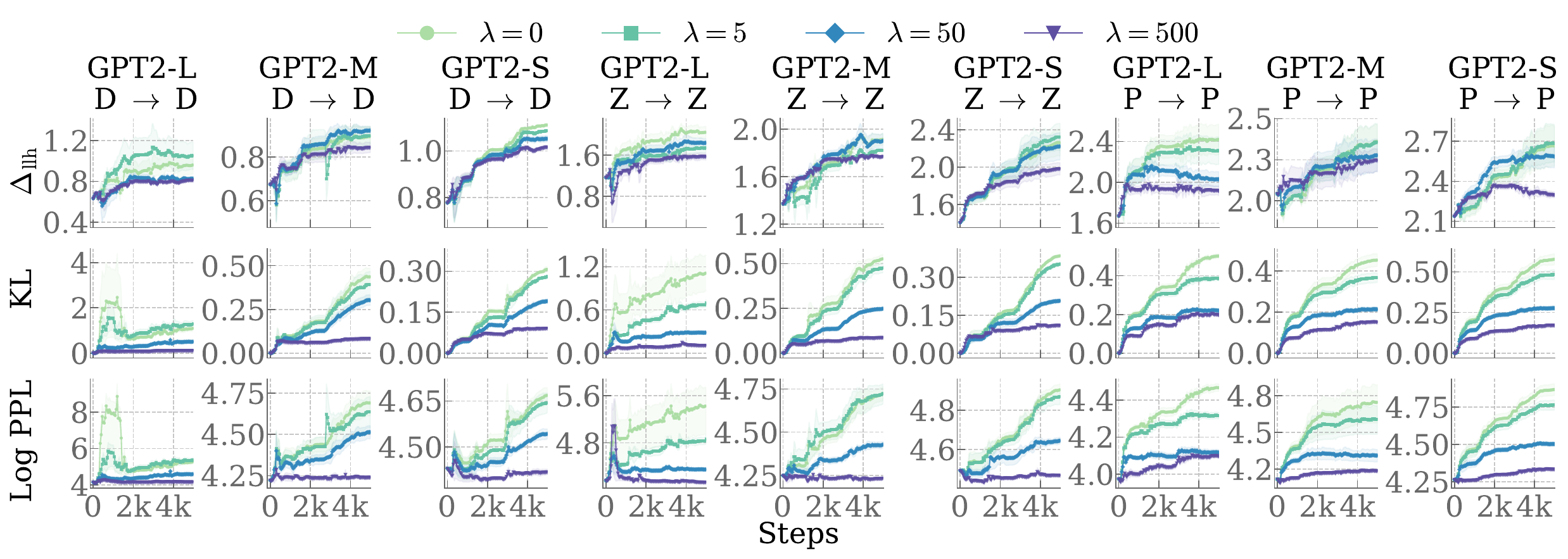}
  \caption {\textbf{Trajectories of $\dll$, KL divergence and log perplexity for KL coefficients $\klReg \in \{0,5,50,500\}$.} Higher coefficients lead to lower perplexity increases as well as lower $\dll$ increases, showing that the KL regularization constrains $\model$ from diverging too much from $\pref$.} 
  \label{fig:kl_comparison}
  \vspace{-10pt}
\end{figure*}
Next, we analyze the effect of adding KL regularization to our objective in \Cref{eqn:loss}.
Specifically, we compare the trajectories of $\dll$, KL divergence and log perplexity for KL coefficients $\klReg \in \{0, 5, 50, 500\}$. 
The coefficients used in this paper are larger than the ones normally used in RLHF studies \citep{schulman-2017-proximal, ziegler-etal-2019-rlhf, stiennon-2020-learning, ouyang-etal-2022-training} because of the large magnitude of our reward in \Cref{eqn:rewardApprox}, which is the negative mean squared error. \Cref{fig:kl_comparison} shows a clear trend: Higher coefficients lead to lower increases in perplexity and KL divergence, as well as lower increases in $\dll$.
We observe higher coefficients reduce the divergence of $\model$ from their initial distribution $\pref$. 
While higher coefficient come at the cost of lower $\dll$ increases, we still observe consistent increases over the baseline; see \Cref{fig:mse_dll_500_app} for results on all data splits with $\klReg=500$.\looseness=-1

\section{Additional Analyses}
Previous work has found that language models fine-tuned on cognitive data, such as eye tracking \citep{yang-etal-2023-plmaspl, deng-etal-2023-pre} and brain data \citep{Toneva2019InterpretingAI} can improve performance on downstream NLP tasks. 
In this section, we evaluate our fine-tuned language models---based on the lowest loss on the test dataset---on several such tasks.
We observe no such improvement.\looseness=-1

\subsection{Targeted Syntactic Evaluation}
\label{ssec:grammar-learning}
\begin{figure}
  \includegraphics[width=\linewidth]{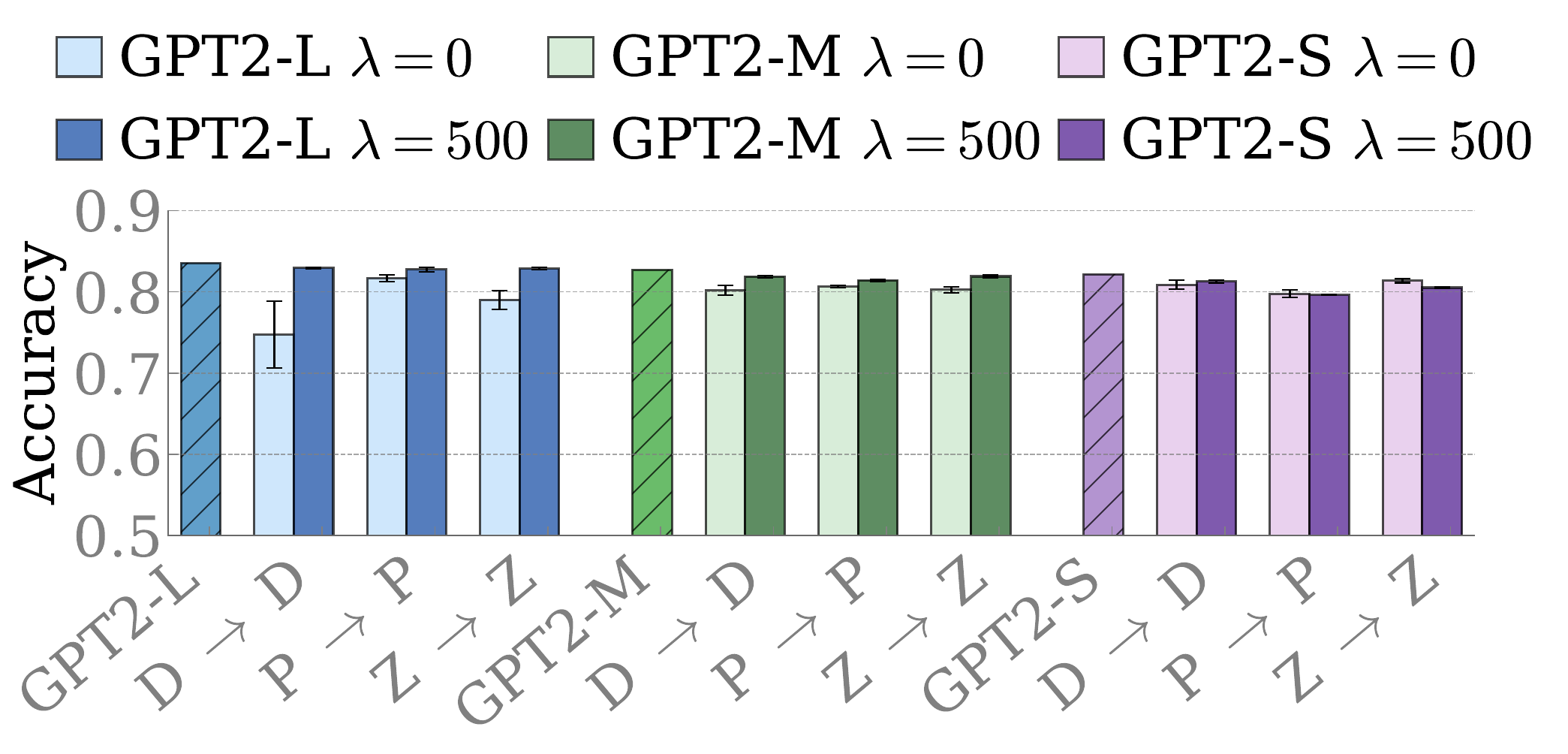}
  \caption{\textbf{Results for BLiMP.}
  Non-fine-tuned models are shown with hatching. 
  Error bars are standard errors across random seeds. Fine-tuning leads to a decrease in accuracy. For results on all data splits, see \Cref{fig:blimp_app}.}
  \label{fig:blimp}
  \vspace{-10pt}
\end{figure}

We evaluate models on the Benchmark of Linguistically Minimal Pairs \citep[BLiMP;][]{warstadt-etal-2020-blimp-benchmark}.\footnote{We use the LM Evaluation Harness \citep{eval-harness}.} 
BLiMP assesses whether language models' behavior is consistent with human preferences for grammatical sentences across a range of grammatical phenomena. In BLiMP, items come in grammatical and ungrammatical variants; we report model accuracy for assigning a higher probability to the grammatical version. Models' scores on BLiMP are visualized in \Cref{fig:blimp}. Even though KL regularization helps mitigate the drop in accuracy, we observe that our fine-tuned models generally exhibit slightly lower accuracy than their non-fine-tuned counterparts, indicating that our fine-tuning procedure does not lead to a better generalization about English grammatical rules in our models. \looseness=-1

\subsection{Text Generation}
\label{ssec:text-gen}
Additionally, we analyze how our objective affects the ability of fine-tuned models to generate text and focus on two aspects: the uniformity of information and the diversity of the generations. To assess uniformity, we draw on the uniform information density (UID) hypothesis \citep{fenk1980, jaeger2006}, which posits that language users prefer information to be evenly distributed throughout an utterance.
A recent study by \citet{meister-etal-2021-revisiting} provides empirical support for the UID hypothesis in naturally occurring corpora and shows a link between linguistic acceptability and information uniformity.
Here, we ask whether aligning models to human reading times encourages them to generate text with greater uniformity of information.\looseness=-1

We test this by generating completions for prefixes from the CNN/DailyMail dataset \citep{hermann-etal-2015-teaching, see-etal-2017-get}. In \Cref{tab:generation_metrics}, we report the mean surprisal variance ($\surprisalvariance$) and unique unigram ratio (1-Gram\%) across all completions; see \Cref{asec:text-generation} for more details.
Models without regularization ($\klReg = 0$) show higher surprisal variance compared to the pretrained models, indicating less uniformly distributed information in the generations. However, under regularization ($\klReg = 500$), this trend is reversed, and we observe lower variance with the exception of GPT2-L: Dundee $\rightarrow$ Dundee. 
We further observe a decrease in the unique unigram ratio, indicating that fine-tuned models generate more repetitive text. 
Overall, these findings suggest that aligning models to human reading times might promote a more uniform information distribution, though further research is needed to explore the connection between the UID hypothesis and alignment with reading times.\looseness=-1

\section{Discussion}
We now discuss the broader implications of our results with respect to both cognitive modeling and natural language processing. 
One recurring theme in this paper is the relationship between $\model$, a probability distribution over strings estimated from large corpora, and $\phuman$, the distribution implicated during cognitive tasks. In terms of cognitive modeling, it is widely accepted that it is useful to obtain a $\model$ that is as close as possible to $\phuman$ in order to test and refine psychological theories. While previous work has proposed alignment procedures between LMs and brain data \citep{Toneva2019InterpretingAI}, within the realm of psycholinguistics, researchers have generally opted for whatever model is the current state of the art. Only recently have contributions started to search for which combinations of training data, model size, and model architecture produce the best cognitive alignment \citep{oh-schuler-2023-transformer, steuer-etal-2023-large}.
Our results suggest an alternative to this search by directly aligning language models to psychometric data. We are optimistic that such aligned models will enable a more precise evaluation of psycholinguistic theories. Of course, it remains an open question to what extent alignment on reading generalizes to other cognitive tasks. We have discussed $\phuman$ in terms of a single hypothesized construct, but it is likely that people's predictions change with task demands, e.g., when skimming versus proofreading a text. 
Testing whether our results hold for other types of psychometric predictive tasks is an important question for future research. \looseness=-1

The main technical contribution of this work is a technique that aligns language models more closely to human psychometric data. Inspired by the implicit parameter optimization in DPO, our approach goes beyond the original Bradley--Terry assumption and demonstrates that it is possible to fine-tune under other implicit statistical models. Exploring a wider range of such models goes hand in hand with exploring new sources of human psychometric data, as we have done here with reading times. Psychologists have devised methods for collecting a diverse array of cognitive signals, including EEG, fMRI, mouse-tracking, and self-paced reading, to name a few. Aligning models, using the method proposed here, on such data or combinations thereof, will be an important next step in this research.\looseness=-1

Finally, what do our results say about the relationship between cognitive modeling and other domains in NLP? At first glance, they seem to suggest that alignment with reading times is not an effective strategy to broadly boost performance on NLP tasks. Although KL regularization reduces the increase of the language model's perplexity on held-out data, we still observe decreased performance on downstream NLP tasks.
However, given the lack of reading time data for LM-generated text, we relied on pre-existing eye-tracking datasets. 
Future studies could experiment with text generated by language models, particularly as new eye-tracking datasets for LM-generated text are being developed \citep{bolliger2024emteccorpuseyemovements}.
Furthermore, \citet{rafailov-2023-direct} derive the optimal solution to the KL regularized RL objective in DPO, while our study excluded the KL term when deriving optimal coefficients. Future work could investigate approaches closer to DPO to compute the optimal coefficients.\looseness=-1

\begin{table}
\small
\centering
\setlength{\tabcolsep}{3.5pt}
\begin{tabular}{lcccc}
\toprule
Model & \multicolumn{2}{c}{$\klReg = 0$} & \multicolumn{2}{c}{$\klReg = 500$} \\
\cmidrule(lr){2-3} \cmidrule(lr){4-5}
 & $\downarrow \surprisalvariance$ & $\uparrow$ 1-Gram\% & $\downarrow \surprisalvariance$ & $\uparrow$ 1-Gram\% \\
\midrule
GPT2-L & $6.69$  & $84.84$ &  $6.69$   & $84.84$ \\
D $\rightarrow$ D & $10.17_{2.93}$  & $67.07_{14.68}$ & $8.05_{0.77}$ & $82.38_{0.18}$ \\
P $\rightarrow$ P & $11.27_{1.74}$  & $68.91_{1.76}$ & $5.14_{0.35}$  & $76.75_{0.79}$ \\
Z $\rightarrow$ Z & $13.85_{0.98}$  & $84.60_{1.91}$ & $5.61_{0.09}$  & $81.12_{0.20}$ \\
\midrule
GPT2-M & $7.67$  & $84.39$ & $7.67$  & $84.39$ \\
D $\rightarrow$ D & $8.94_{0.25}$  & $80.82_{0.35}$ & $6.63_{0.35}$ & $83.22_{1.08}$ \\
P $\rightarrow$ P & $7.81_{0.46}$  & $67.51_{0.78}$ & $4.54_{0.17}$ & $75.91_{0.85}$ \\
Z $\rightarrow$ Z & $12.12_{1.42}$ & $84.95_{0.86}$ & $5.88_{0.05}$ & $81.94_{0.91}$ \\
\midrule
GPT2-S & $5.70$  & $82.53$ & $5.70$ & $82.53$ \\
D $\rightarrow$ D & $6.27_{0.52}$  & $81.14_{0.95}$ & $4.50_{0.11}$  & $76.56_{2.02}$ \\
P $\rightarrow$ P & $6.91_{0.25}$ & $72.17_{1.99}$ & $4.11_{0.10}$ & $69.54_{0.43}$ \\
Z $\rightarrow$ Z & $7.79_{1.02}$  & $79.61_{1.94}$ & $4.97_{0.18}$  & $75.92_{0.68}$ \\
\bottomrule
\end{tabular}
\caption{Surprisal variance ($\surprisalvariance$), and unique unigram ratios (1-Gram\%) of model generated completions. See \Cref{tab:generation_metrics_app} for results on all data splits.}
\label{tab:generation_metrics}
\vspace{-12pt}
\end{table}

\section{Conclusion}
We have presented a novel technique to align language models with human reading data by implicitly optimizing the parameters of a linear regression model. Our experiments on held-out test sets show our method reliably improves the predictive power of language models with various parameter counts on human reading times. Furthermore, our findings confirm previous research on the inverse relationship between perplexity and psychometric predictive power. We believe our results pave the way for better assessment of psychological theories using more cognitively aligned language models.\looseness=-1

\section*{Limitations}

Our study has several limitations. First, we only include predictors for the current unit. Future research could explore the impact of adding previous units' predictors as well as additional predictors, such as contextual entropy, as suggested in \citet{pimentel-etal-2023}, to the objective. Second, our study only tests and evaluates English language data. Expanding these studies to a wider variety of languages is an important next step in establishing the generalizability of our methods. Third, we estimate surprisal based on individual sentences, whereas surprisal estimated for whole paragraphs may yield more accurate estimates due to the additional context.\looseness=-1

\section*{Ethics Statement}

Our study introduces a technique for aligning language models to human psychometric data. When working with human psychometric data, specifically eye-tracking data, it is important to consider potential privacy risks \citep{jaeger-etal-2020-deep, lohr-etal-2022-eye}. In this study, we used well-established datasets \citep{kennedy-etal-2003-dundee, luke-etal-2018-provo, hollenstein-etal-2018-zuco1, hollenstein-etal-2020-zuco2}, where personal identifying information had been anonymized prior to our access.
Additionally, we are aware of the potential biases inherent in language models and human reading data. Our goal is to ensure that our models and evaluations do not propagate or amplify existing biases.

\section*{Acknowledgments}
We would like to thank the ARR reviewers and area chair for their valuable feedback and comments.
Afra Amini is supported by ETH AI Center doctoral fellowship. David Robert Reich is supported by the Swiss National Science Foundation (SNSF) under grant IZCOZ0\_220330/1 (EyeNLG).

\bibliography{custom}
\newpage
\appendix
\onecolumn
\label{asec:appendix}

\section{Derivations}
\subsection{Deriving Optimal Coefficients} \label{asec:regression}
We now discuss how we estimate the optimal coefficient $\bbeta^{\star}$ to approximate the reward, as described in \Cref{ssec:reward-approximation}. We start with finding the optimal coefficients
\begin{equation}
    \betaRegTrg^\star = \argmin_{\bbeta\in\R^D}\frac{1}{N} \|\bpsychometric - \featmat \betaRegTrg \|^2.
\end{equation}
Assuming $\big(\featmat^{\top}\featmat\big)^{-1}$ is invertible, we take the derivative with respect to $\betaRegTrg$, set it to 0 and solve for $\betaRegTrg^{\star}$:
\begin{subequations}
   \begin{alignat}{2}
    & \quad -\frac{2}{N}\featmat^{\top}\big(\bpsychometric - \featmat \betaRegTrg^{\star}\big) & & = \quad0 \\
    \Leftrightarrow & \quad -\featmat^{\top}\bpsychometric + \featmat^{\top}\featmat \betaRegTrg^{\star} & & =\quad 0 \\
    \Leftrightarrow & \quad\quad \featmat^{\top}\featmat \betaRegTrg^{\star} & & = \quad\featmat^{\top}\bpsychometric \\
    \Leftrightarrow & \quad\quad \betaRegTrg^{\star} & & = \quad\big(\featmat^{\top}\featmat\big)^{-1} \featmat^{\top} \bpsychometric.
\end{alignat}
\end{subequations}
In theory, $\big(\featmat^{\top}\featmat\big)^{-1}$ may not always be invertible, which is why we add a regularization term $\idscale \id$, where $\id$ is the identity matrix and $\idscale > 0 $ is a parameter determining the strength of the regularization. The resulting estimator $\betaRegTrg^{\star} = \big(\featmat^{\top}\featmat + \idscale \id\big)^{-1} \featmat^{\top} \bpsychometric$ is known as the ridge regression estimator \citep{hoerl-1970-ridge} and presents the solution to the following problem:
\begin{equation}
\label{eqn:ridge-obj}
   \betaRegTrg^\star = \argmin_{\betaRegTrg\in\R^D}\frac{1}{N} \|\bpsychometric - \featmat \betaRegTrg \|^2 + \ridgeReg\|\betaRegTrg\|^2,
\end{equation}
where $\ridgeReg >0$. We define $\idscale = N\ridgeReg$. Then setting the derivative of \Cref{eqn:ridge-obj} to zero leads to 
\begin{subequations}
   \begin{alignat}{2}
    & \quad -\frac{2}{N}\featmat^{\top}\big(\bpsychometric - \featmat \betaRegTrg^{\star}\big)+ 2\ridgeReg\betaRegTrg^{\star} & & = \quad0 \\
    \Leftrightarrow & \quad\quad \frac{2}{N}\featmat^{\top}\featmat \betaRegTrg^{\star} + 2\ridgeReg\betaRegTrg^{\star} & & = \quad\frac{2}{N}\featmat^{\top}\bpsychometric \\
    \Leftrightarrow & \quad\quad \featmat^{\top}\featmat \betaRegTrg^{\star} + N\ridgeReg\betaRegTrg^{\star} & & = \quad\featmat^{\top}\bpsychometric \\
    \Leftrightarrow & \quad\quad \featmat^{\top}\featmat \betaRegTrg^{\star} + \idscale\betaRegTrg^{\star} & & = \quad\featmat^{\top}\bpsychometric \\
    \Leftrightarrow & \quad\quad \betaRegTrg^{\star} & & = \quad\big(\featmat^{\top}\featmat + \idscale \id \big)^{-1} \featmat^{\top} \bpsychometric.
\end{alignat}
\end{subequations}

\subsection{Solving for Optimal Coefficients}
\label{asec:cholesky}
To compute the regression coefficients efficiently, we use the Cholesky decomposition to solve for $\betaRegTrg^\star$ given as
\begin{align}
\betaRegTrg^{\star} = (\featmat^{\top} \featmat + \idscale \id)^{-1} \featmat^{\top} \bpsychometric,
\end{align}
where $\idscale$ is the regularization parameter, which we set $\idscale=1\mathrm{e}-5$ and $\id$ is the identity matrix. Since $\featmat^{\top} \featmat + \idscale \id$ is symmetric and positive definite, we compute the Cholesky decomposition
\begin{align}
\featmat^{\top} \featmat + \idscale \id = \choleskyL\choleskyL^{\top},
\end{align}
where $\choleskyL \in \R^{D\times D}$ is a lower triangular matrix. To solve for $\betaRegTrg^{\star}$, we first solve for an intermediate vector $\choleskyz$
 \begin{align}
\choleskyL\choleskyz = \featmat^{\top} \bpsychometric
\end{align}
We then solve for $\betaRegTrg^{\star}$
\begin{align}
\choleskyL^{\top} \betaRegTrg^{\star} = \choleskyz.
\end{align}
\newpage

\section{Datasets, Reading Times \& Parameters}
\subsection{Datasets}
\label{asec:datasets}
Here, we provide additional details on the datasets and reading time measurements used during our analysis. We fine-tune and evaluate models on the Dundee \cite{kennedy-etal-2003-dundee}, Provo \cite{luke-etal-2018-provo}, and ZuCo corpora. For the ZuCo corpus, we use data from tasks 1 and 2 from the ZuCo 1.0 corpus \cite{hollenstein-etal-2018-zuco1} and task 1 from the ZuCo 2.0 corpus \citep{hollenstein-etal-2020-zuco2}. All data used in our analysis is publicly available. For the Dundee and ZuCo corpora, we process the data used by \citet{hollenstein-etal-2021-multilingual}, which contains word-level means for fixation counts and reading durations, averaged over all participants, and split into individual sentences.\footnote{\url{https://github.com/DS3Lab/multilingual-gaze}} For the Provo corpus, we compute the mean reading times from the official repository. \footnote{\url{https://osf.io/sjefs/}} From all datasets, we remove duplicate sentences and short sentences with less than four words.
The mean number of sentences and words for train and test sets are given in \Cref{tab:data_splits_configurations_app}.
\begin{table}[htbp!]
\small
\centering
\setlength{\tabcolsep}{3.5pt}
\begin{tabular}{lcccc} 
\toprule Configuration & Train Tokens & Train Sents & Test Tokens & Test Sents \\ 
\midrule 
Dundee (D) $\rightarrow$ Provo (P) & 20894.7 & 980 & 1144 & 54  \\ 
Dundee (D) $\rightarrow$ Dundee (D) & 20894.7 & 980  & 20207 & 931 \\ 
Dundee (D) $\rightarrow$ ZuCo (Z) & 20894.7 & 980 & 7715 & 424  \\ 
\midrule 
ZuCo (Z) $\rightarrow$ Provo (P) & 7761 & 451 & 1144 & 54  \\ 
ZuCo (Z) $\rightarrow$ Dundee (D) & 7761 & 451&  20207 & 931 \\ 
ZuCo (Z) $\rightarrow$ ZuCo (Z) & 7761 & 451 & 7715 & 424  \\ 
\midrule 
Provo (P) $\rightarrow$ Provo (P) & 1113.7 & 56 & 1144 & 54 \\ 
Provo (P) $\rightarrow$ Dundee (D) & 1113.7 & 56 & 20207 & 931  \\ 
Provo (P) $\rightarrow$ ZuCo (Z) & 1113.7 & 56 & 7715 & 424 \\ 
\bottomrule 
\end{tabular} 
\caption{Data splits and configurations for fine-tuning and evaluation. Numbers indicate the \textbf{mean} number of tokens and sentences in each train and test split across random seeds.} \label{tab:data_splits_configurations_app} 
\end{table}
\subsection{Reading Times}
For Dundee and ZuCo, we extract the mean first pass duration over the participants, which is defined as
``the sum of all fixations on w from the first time a subject fixates w to the first time the subject fixates another token''\citep[p.109]{hollenstein-etal-2021-multilingual}. Similarly, for Provo, we compute the mean gaze duration defined as the ``Dwell time (i.e., summation of the duration across all fixations) of the first run within the current interest area'' \citep[Tab. 2]{luke-etal-2018-provo}. While these two definitions are very similar, they may not be exactly identical.
In \Cref{tab:app_dataset_comparison}, we compare the mean and standard deviation of reading times as well as the number of zero reading times. We observe that while Dundee and Provo exhibit relatively similar means and standard deviations, ZuCo shows overall lower mean reading times. Unlike Dundee and ZuCo, Provo contains no instances of words with zero reading times, likely due to the high number of participants.

\begin{table}[h!]
\small
\centering
\begin{tabular}{lrrr}
    \toprule
    Dataset       & Mean Reading Times & STD Reading Times &  Zero Count  \\
    \midrule
    Dundee        & 140.59 & 88.49 & 854 \\
    Provo         & 164.16 & 77.77 & 0  \\
    ZuCo          & 92.28 & 52.21 & 176 \\
    \bottomrule
\end{tabular}
\caption{Mean, standard deviation, and zeros count for the reading times from Dundee, Provo, and ZuCo.}
\label{tab:app_dataset_comparison}
\end{table}

\subsection{Fine-Tuning Parameters}
\label{asec:trainingparameters}
For all runs, we use the parameter configurations in \Cref{tab:training-parameters} and repeat each experiment 3 times with random seeds (42, 8, and 64). We fine-tune and evaluate three GPT-2 models: GPT-2 Small with 117 million parameters, GPT-2 Medium with 345 million parameters, and GPT-2 Large with 762 million parameters. All models are fine-tuned for 5k data steps using a batch size of 1 and gradient accumulation of 2, leading to a total of 2.5k optimization steps.
To adjust the learning rate during fine-tuning, we use a cosine annealing learning rate schedule\footnote{\url{https://github.com/katsura-jp/pytorch-cosine-annealing-with-warmup}} \citep{loshchilov-2017-sgdr} with a maximum learning rate of 1.5e-5, minimum learning rate of 2e-7. During each cycle, we decrease the learning by a factor of 0.8 and increase the cycle length by a factor of 1.8.
For all experiments, we use NVIDIA GeForce RTX 3090, RTX 4090, and RTX 2080 Ti GPUs. The GPU times vary depending on the model's parameter counts and the size of the data splits, taking a minimum of approximately 12 minutes for GPT-2 Small fine-tuned on Provo and evaluated on Provo and a maximum of 4 hours for GPT-2 Large fine-tuned on Dundee and evaluated on Dundee.

\begin{table}[h!]
\small
\centering
\begin{tabular}{ll}
\toprule
Parameter & Setting  \\ 
\midrule
Optimizer & AdamW \\
Scheduler & Cosine Annealing With Warm Restarts  \\
Batch Size & 1 \\
Grad. Accumulation & 2 \\
Total Steps & 5000 \\
Optimizer Steps & 2500 \\
Max Learning Rate & 1.5e-5 \\
Min Learning Rate & 2.0e-7 \\
Decrease Rate of Max Learning Rate & 0.8 \\
Cycle Steps Magnification & 1.8 \\
Warm Up Steps & 100 \\
\bottomrule
\end{tabular}
\caption{Fine-tuning parameters used consistently across different runs.}
\label{tab:training-parameters}
\end{table}

\subsection{Evaluation}
We perform evaluation every 50 steps on the held-out test splits in \Cref{tab:data_splits_configurations_app}, during which we compute the surprisal estimates, the regressors' coefficients, and the language model's perplexity for each batch. 
Using the surprisal estimates we compute the $\dll$ by performing a 5-fold cross-validation on the test data, where we fit baseline and target linear regressors using ordinary least squares.\footnote{\url{https://www.statsmodels.org/stable/generated/statsmodels.regression.linear_model.OLS.html}} Note that we do not scale the predictor variables to maintain consistency between the linear regression and the calculation of our reward in \Cref{eqn:rewardApprox}, where batch-level surprisal estimates prevent global scaling.
\newpage

\section{Detailed Results}
\subsection{$\dll$ Change}
\begin{table}[h!] 
    \small
    \centering
    \begin{tabular}{l|ccc|ccc|ccc} 
    \toprule 
    \multicolumn{1}{c}{} & \multicolumn{3}{c}{GPT2-L} & \multicolumn{3}{c}{GPT2-M} & \multicolumn{3}{c}{GPT2-S} \\
    Data &  $\dll^{\text{start}}$ &  $\dll^{\text{max}}$ & \% Increase 
              &  $\dll^{\text{start}}$ &  $\dll^{\text{max}}$ & \% Increase 
              &  $\dll^{\text{start}}$ &  $\dll^{\text{max}}$ & \% Increase \\ 
     \midrule  
    D $\rightarrow$ D & $0.63$ & $0.98 \pm 0.09$ & 55.29 & $0.68$ & $0.90 \pm 0.06$ & 32.82 & $0.78$ & $1.11 \pm 0.01$ & 42.88 \\ 
    P $\rightarrow$ D & $0.63$ & $0.78 \pm 0.04$ & 23.29 & $0.68$ & $0.84 \pm 0.00$ & 24.23 & $0.78$ & $1.01 \pm 0.02$ & 29.90 \\ 
    Z $\rightarrow$ D & $0.63$ & $0.69 \pm 0.01$ & 9.08  & $0.68$ & $0.75 \pm 0.03$ & 11.08 & $0.78$ & $0.91 \pm 0.04$ & 16.60 \\ 
    D $\rightarrow$ P & $1.67$ & $2.19 \pm 0.37$ & 31.36 & $2.04$ & $2.14 \pm 0.05$ & 4.93  & $2.14$ & $2.52 \pm 0.09$ & 17.78 \\ 
    P $\rightarrow$ P & $1.67$ & $2.43 \pm 0.14$ & 45.27 & $2.04$ & $2.36 \pm 0.10$ & 15.62 & $2.14$ & $2.67 \pm 0.15$ & 24.77 \\ 
    Z $\rightarrow$ P & $1.67$ & $1.71 \pm 0.03$ & 2.12  & $2.04$ & $2.18 \pm 0.03$ & 6.46  & $2.14$ & $2.18 \pm 0.01$ & 2.00  \\ 
    D $\rightarrow$ Z & $1.17$ & $1.59 \pm 0.15$ & 36.58 & $1.37$ & $1.53 \pm 0.06$ & 11.95 & $1.41$ & $1.75 \pm 0.06$ & 24.50 \\ 
    P $\rightarrow$ Z & $1.17$ & $1.34 \pm 0.09$ & 15.01 & $1.37$ & $1.45 \pm 0.08$ & 5.73  & $1.41$ & $1.69 \pm 0.07$ & 20.09 \\ 
    Z $\rightarrow$ Z & $1.17$ & $2.12 \pm 0.10$ & 81.39 & $1.37$ & $1.94 \pm 0.04$ & 41.80 & $1.41$ & $2.26 \pm 0.14$ & 61.20 \\  
    \bottomrule
    \end{tabular}
    \caption{Mean start and maximum $\dll (10^{-2} \text{ nats})$ for \Cref{tab:dll-table}, including standard errors across random seeds, rounded to two decimal places.}
    \label{tab:dll-table-app}
\end{table}

\subsection{Perplexity and $\dll$}
\begin{figure*}[h!]
  \includegraphics[width=0.33\linewidth]{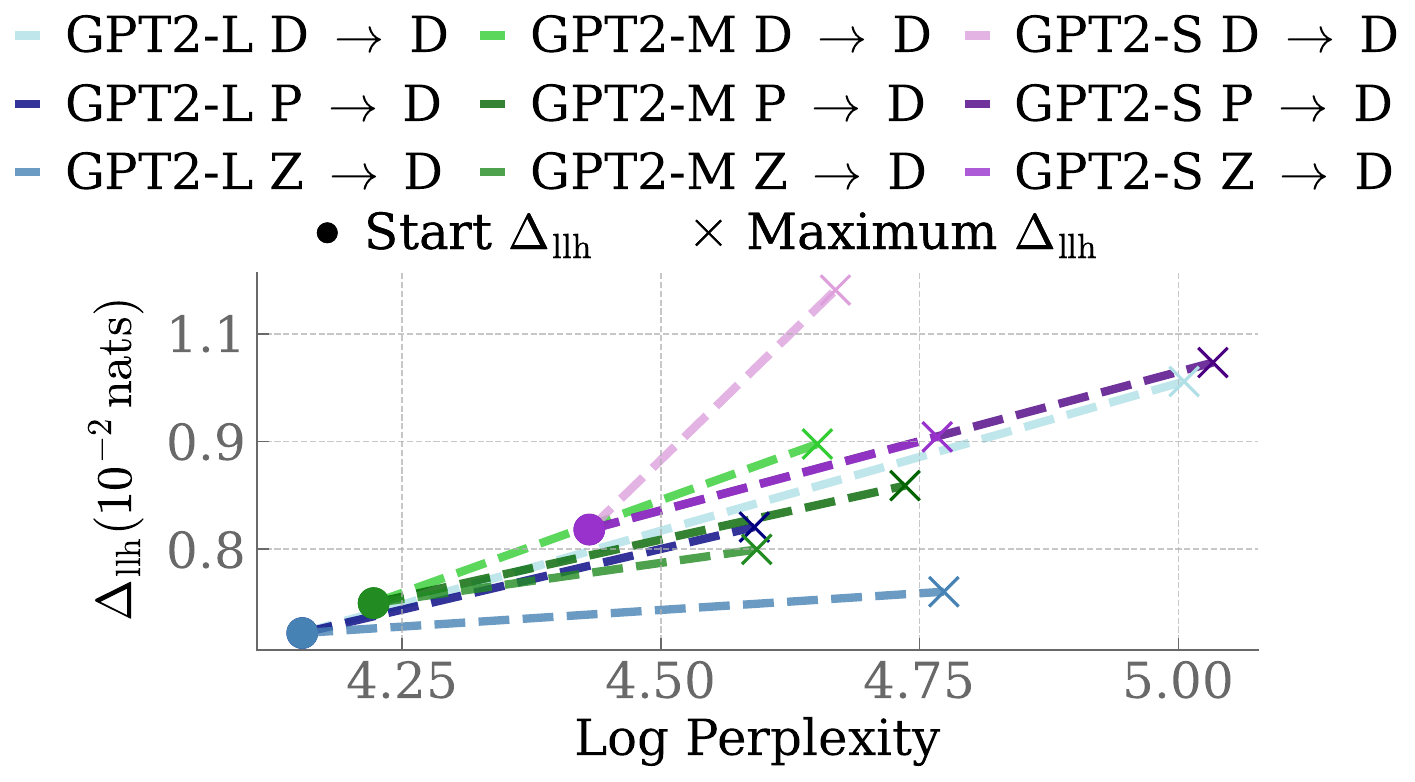}
  \includegraphics[width=0.33\linewidth]{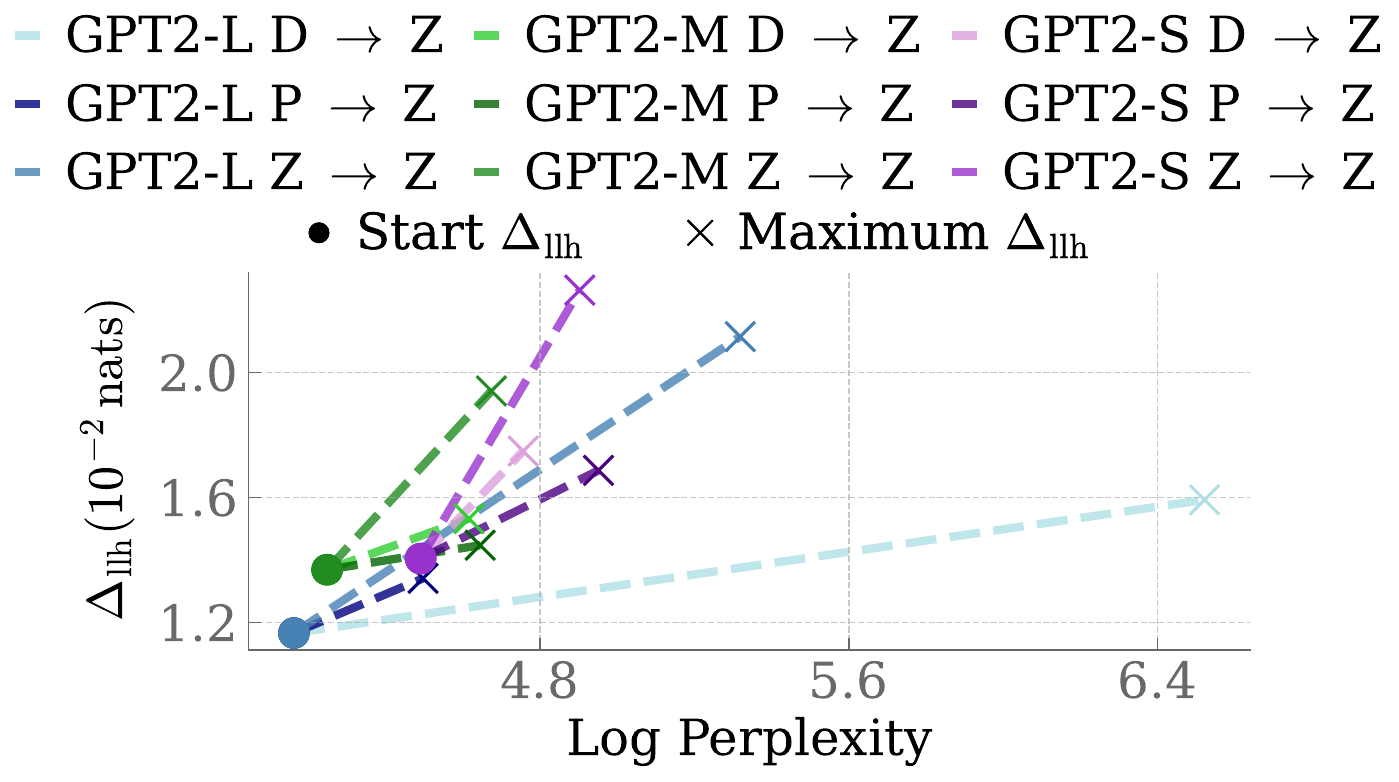}
  \includegraphics[width=0.33\linewidth]{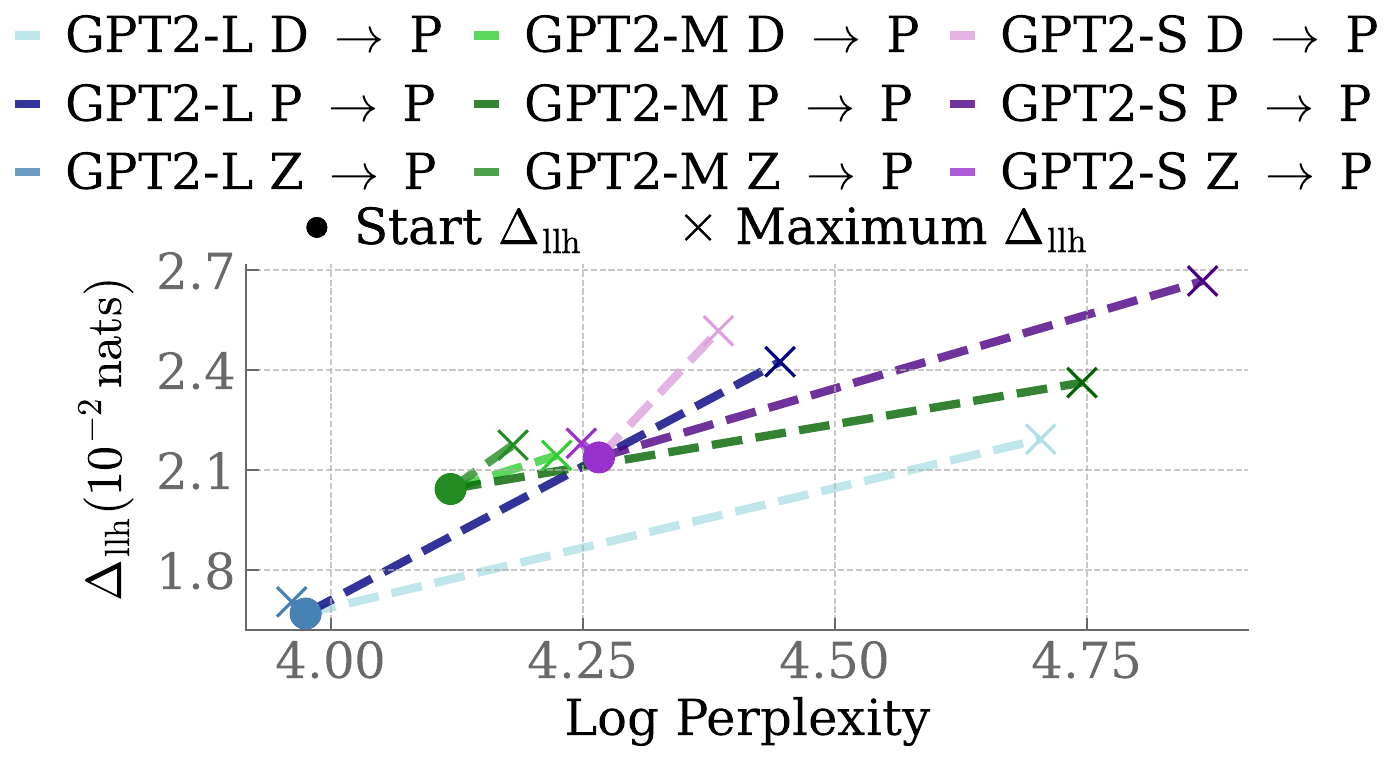}
  \caption {Relationship between log perplexity and $\dll (10^{-2} \text{ nats})$ for all models and data splits. Increases in $\dll$ correspond to increases in perplexity.}
  \label{fig:dll_ppl_app}
\end{figure*}

\subsection{BLiMP}
\label{asec:blimp_app} 
\begin{figure*}[h!]
  \includegraphics[width=0.33\linewidth]{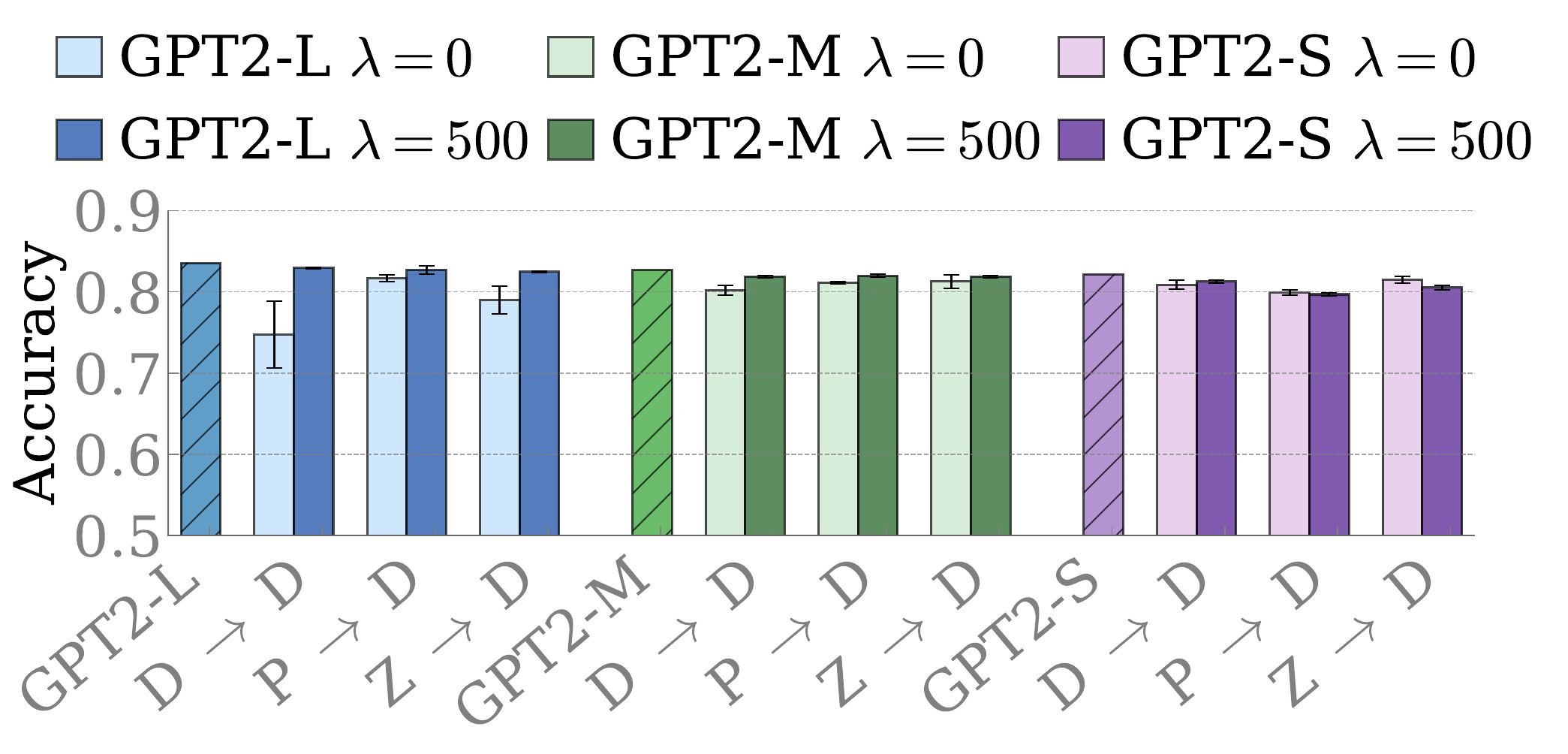}
  \includegraphics[width=0.33\linewidth]{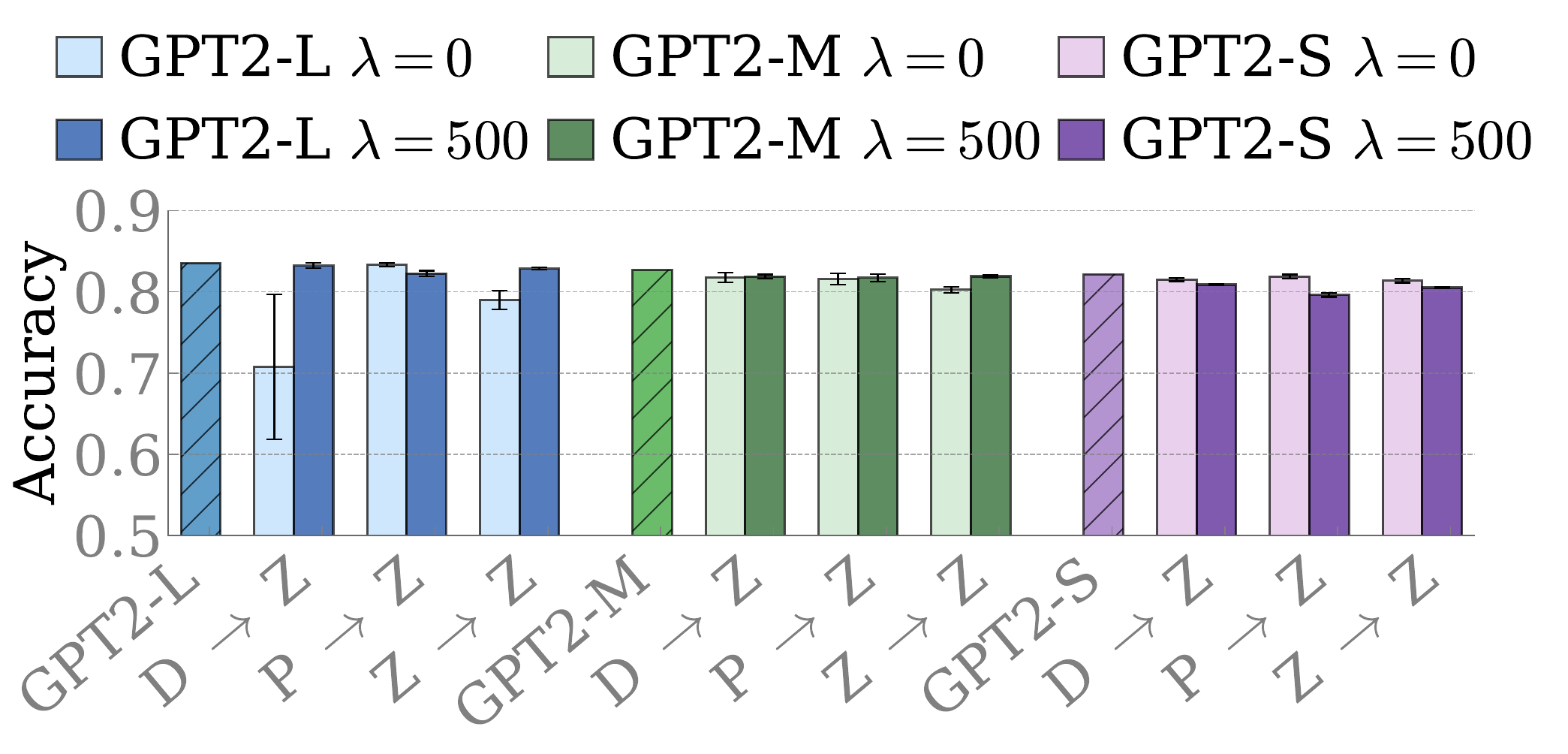}
  \includegraphics[width=0.33\linewidth]{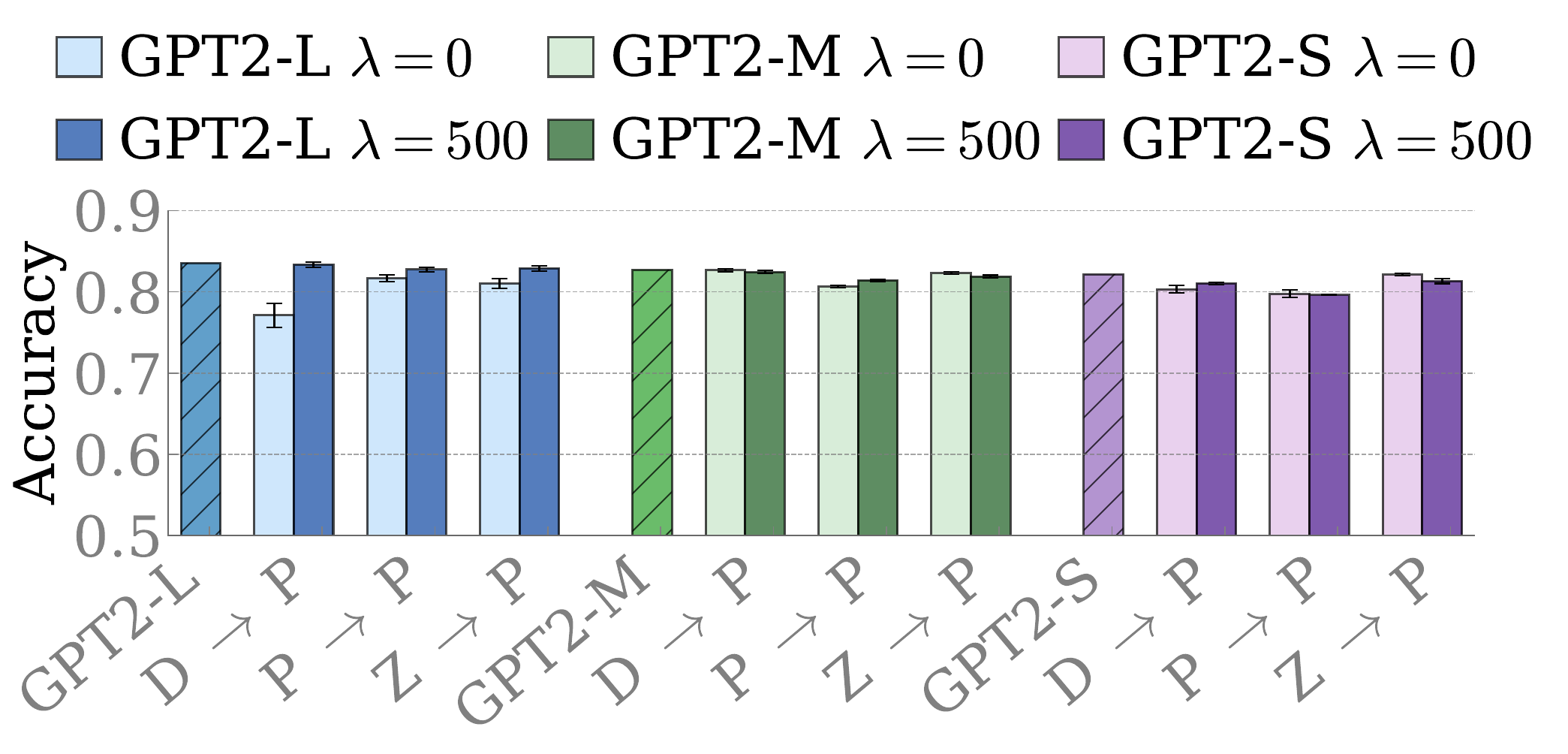}
  \caption {Results for all models on BLiMP. Baseline (i.e., non-fine-tuned) models are shown with hatching. Error bars are
standard errors across the three random seeds. We observe a drop in accuracy as a result of fine-tuning.}
  \label{fig:blimp_app}
\end{figure*}

\subsection{Narrative Understanding}
\begin{figure}[h!]
  \includegraphics[width=0.33\linewidth]{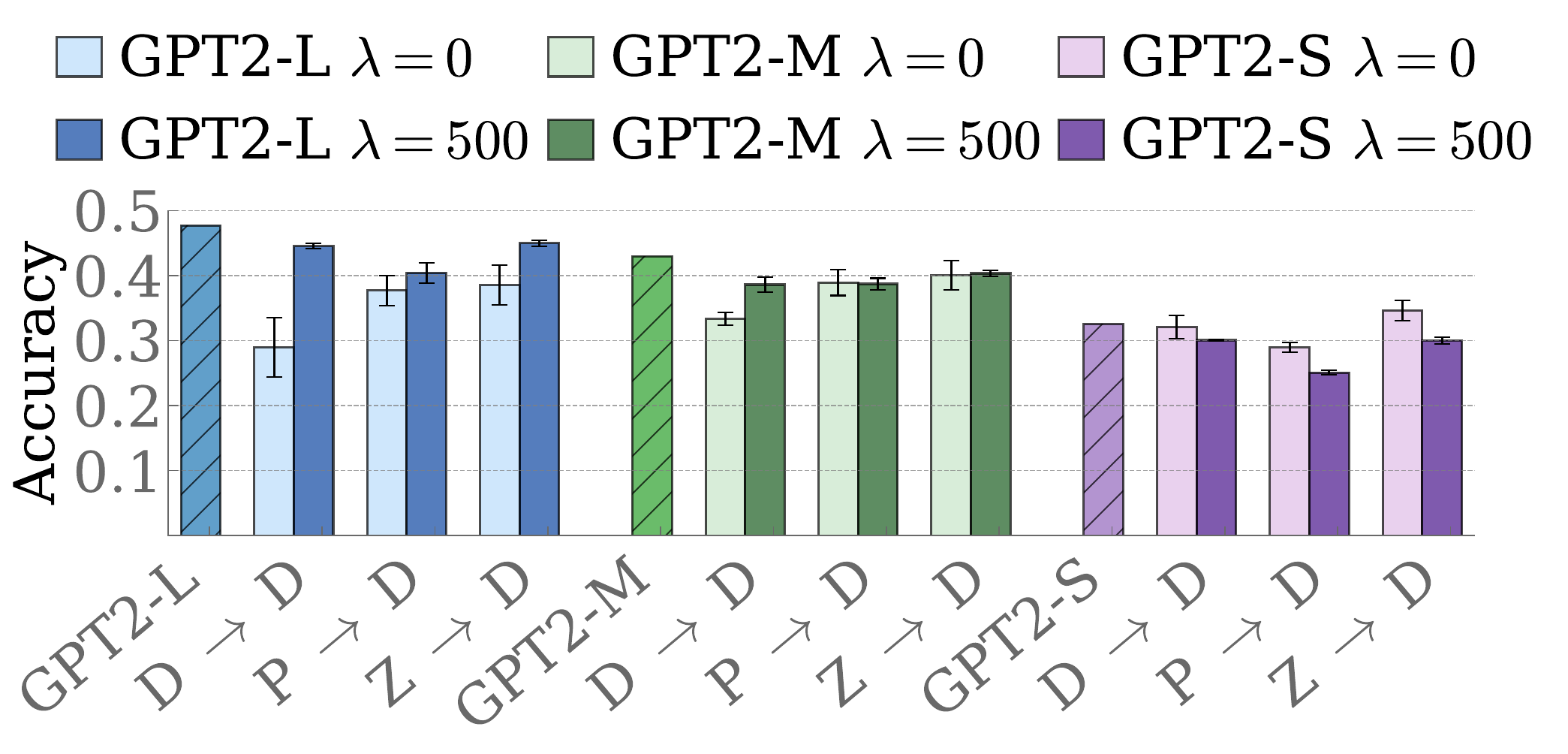}
  \includegraphics[width=0.33\linewidth]{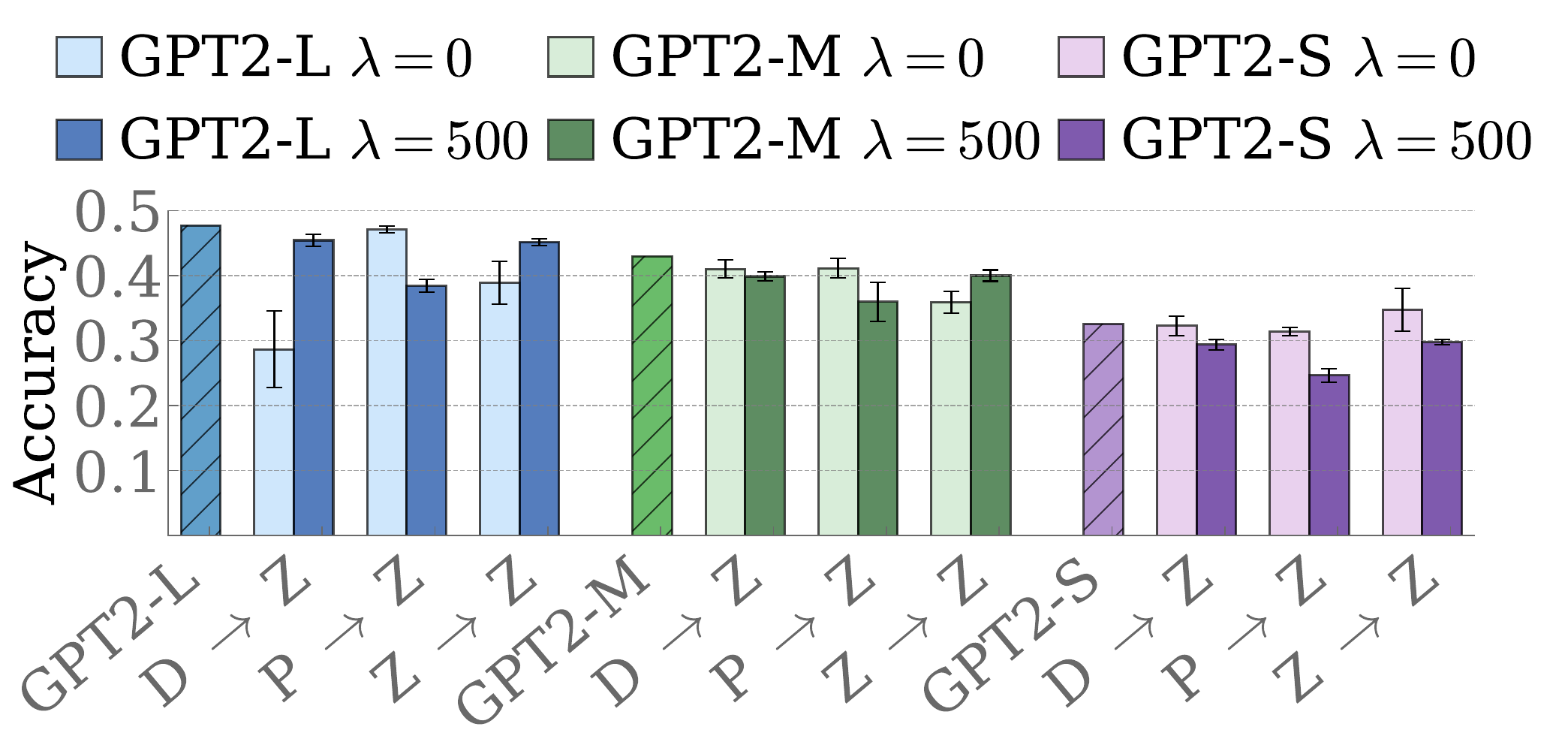}
  \includegraphics[width=0.33\linewidth]{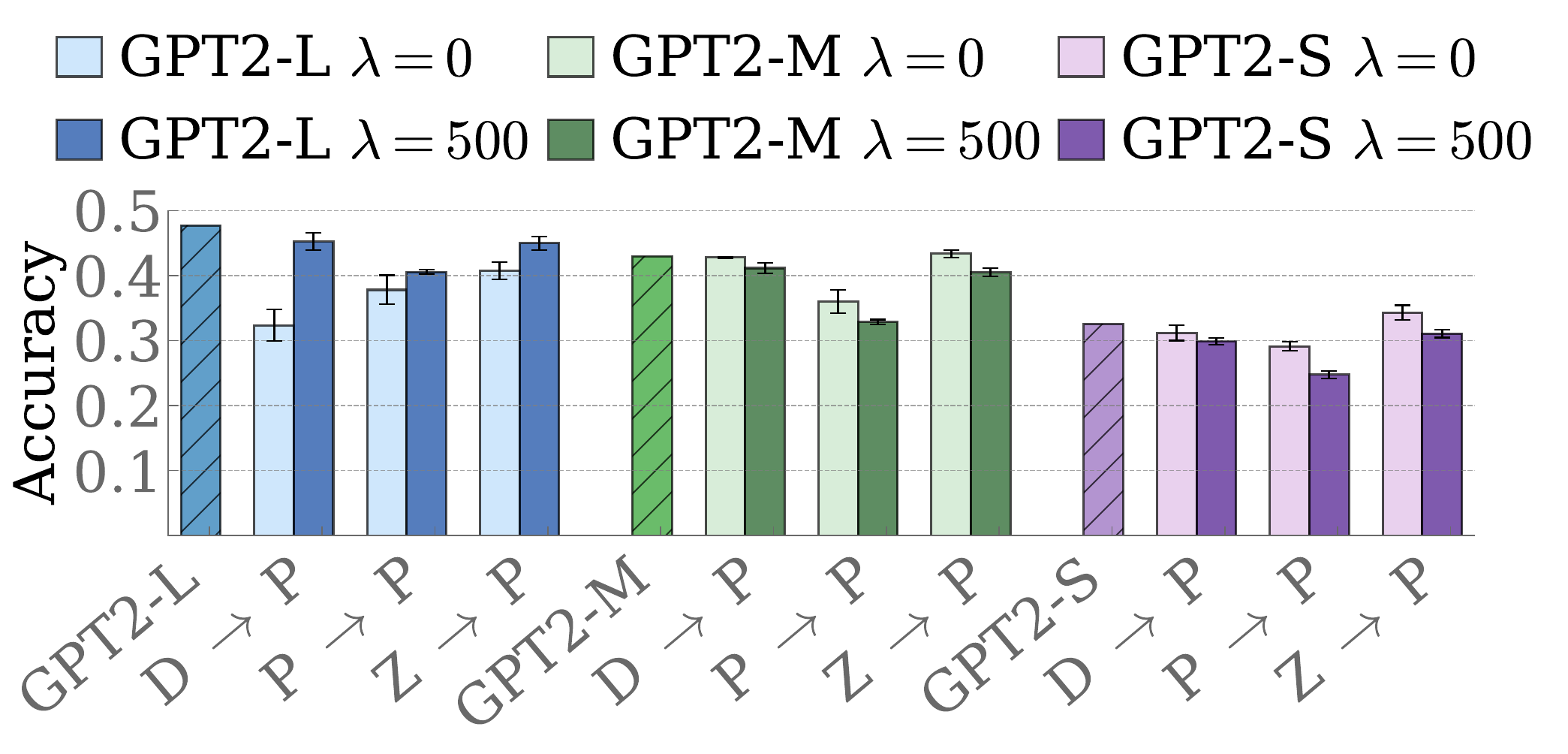}
  \caption{\textbf{Results for LAMBADA.} Baseline (i.e., non-fine-tuned) models are shown with hatching. Error bars are standard errors across the three random seeds. As with BLiMP, performance tends to decrease with fine-tuning.}
  \label{fig:lambada}
\end{figure}

To measure models' abilities to track entities and produce text that is consistent with narrative structure, we evaluate them on LAMBADA \cite{paperno-etal-2016-lambada}. This dataset requires that models produce the final word of a narrative. People can easily achieve higher performance on this task if they are given the full narrative context, but not if they are only given the previous sentence. Thus, performing well requires an understanding of broader contexts. The performance of our models is visualized in \Cref{fig:lambada}. As with BLiMP, we find that fine-tuned models perform slightly worse at this task compared to non-fine-tuned baselines.

\subsection{Results for the Regularized Objective}

\begin{figure*}[h!]
  \includegraphics[width=\linewidth]{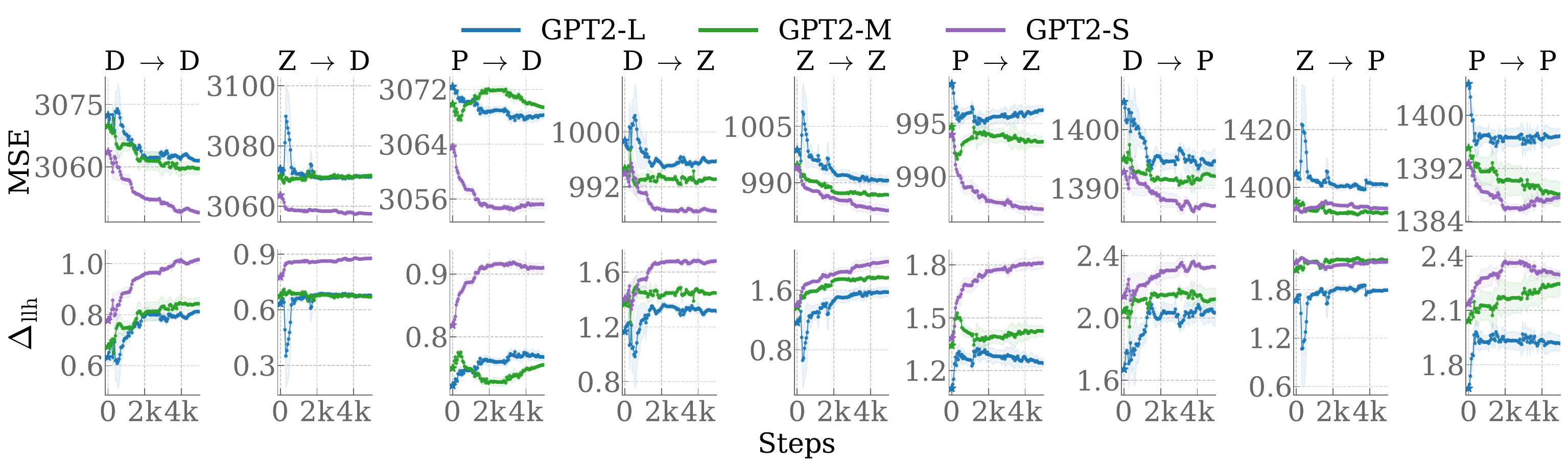}
  \caption {MSE and $\dll (10^{-2} \text{ nats})$ changes for all data splits using the \textbf{KL regularized objective} with $\klReg = 500$. Bands show the standard error across seeds. Compared to the results for the unregularized objective in \Cref{ssec:predicting_rts}, we observe smaller but consistent decreases in MSE and increases in $\dll$ for almost all configurations.}
  \label{fig:mse_dll_500_app}
\end{figure*}

In this section, we present additional results for fine-tuning models using the KL regularized objective with a regularization coefficient of $\klReg = 500$. 
As shown in \Cref{tab:500-dll-table-app}, the maximum $\dll$ values and percentage increases tend to be lower compared to those for the unregularized objective in \Cref{tab:dll-table-app}. However, standard errors for the maximum $\dll$ are consistently lower, suggesting that fine-tuning is more stable across random seeds when regularizing the reward. Additionally, when visualizing the MSE and $\dll$ throughout fine-tuning (\Cref{fig:mse_dll_500_app}), we observe more consistent improvements for Dundee $\rightarrow$ Provo and ZuCo $\rightarrow$ Provo compared to the trajectories for the unregularized objective in \Cref{fig:mse-dll-change}.

\begin{table}[h!] 
    \small
    \centering
    \begin{tabular}{l|ccc|ccc|ccc} 
    \toprule 
    \multicolumn{1}{c}{} & \multicolumn{3}{c}{GPT2-L} & \multicolumn{3}{c}{GPT2-M} & \multicolumn{3}{c}{GPT2-S} \\
    Data &  $\dll^{\text{start}}$ &  $\dll^{\text{max}}$ & \% Increase 
              &  $\dll^{\text{start}}$ &  $\dll^{\text{max}}$ & \% Increase 
              &  $\dll^{\text{start}}$ &  $\dll^{\text{max}}$ & \% Increase \\ 
     \midrule  
    D $\rightarrow$ D & $0.63$ & $0.81 \pm 0.00$ & 28.27 & $0.68$ & $0.85 \pm 0.03$ & 25.49 & $0.78$ & $1.02 \pm 0.01$ & 30.72 \\ 
    P $\rightarrow$ D & $0.63$ & $0.71 \pm 0.01$ & 12.69 & $0.68$ & $0.71 \pm 0.00$ & 5.62 & $0.78$ & $0.93 \pm 0.01$ & 19.34 \\ 
    Z $\rightarrow$ D & $0.63$ & $0.69 \pm 0.00$ & 8.63 & $0.68$ & $0.71 \pm 0.02$ & 5.24 & $0.78$ & $0.88 \pm 0.00$ & 13.22 \\ 
    D $\rightarrow$ P & $1.67$ & $2.09 \pm 0.03$ & 24.91 & $2.04$ & $2.18 \pm 0.03$ & 6.87 & $2.14$ & $2.36 \pm 0.03$ & 10.47 \\ 
    P $\rightarrow$ P & $1.67$ & $1.98 \pm 0.04$ & 18.70 & $2.04$ & $2.25 \pm 0.07$ & 10.00 & $2.14$ & $2.37 \pm 0.02$ & 10.95 \\ 
    Z $\rightarrow$ P & $1.67$ & $1.86 \pm 0.03$ & 11.13 & $2.04$ & $2.19 \pm 0.00$ & 7.38 & $2.14$ & $2.19 \pm 0.01$ & 2.58 \\ 
    D $\rightarrow$ Z & $1.17$ & $1.36 \pm 0.03$ & 16.86 & $1.37$ & $1.50 \pm 0.07$ & 9.52 & $1.40$ & $1.69 \pm 0.02$ & 20.01 \\ 
    P $\rightarrow$ Z & $1.17$ & $1.35 \pm 0.04$ & 16.14 & $1.37$ & $1.52 \pm 0.01$ & 11.24 & $1.40$ & $1.76 \pm 0.03$ & 25.34 \\ 
    Z $\rightarrow$ Z & $1.17$ & $1.58 \pm 0.07$ & 35.24 & $1.37$ & $1.78 \pm 0.01$ & 29.88 & $1.40$ & $1.98 \pm 0.06$ & 41.12 \\ 
    \bottomrule
    \end{tabular}
    \caption{Mean start and maximum $\dll (10^{-2} \text{ nats})$ values using the \textbf{KL regularized objective} with $\klReg = 500$, including standard errors across random seeds, rounded to two decimal places.}
    \label{tab:500-dll-table-app}
\end{table}

\clearpage

\section{Reading Times}
\subsection{Total Reading Duration \& First Fixation Duration}
\label{asec:trt-ffd}
\begin{figure*}[h!]
  \includegraphics[width=\linewidth]{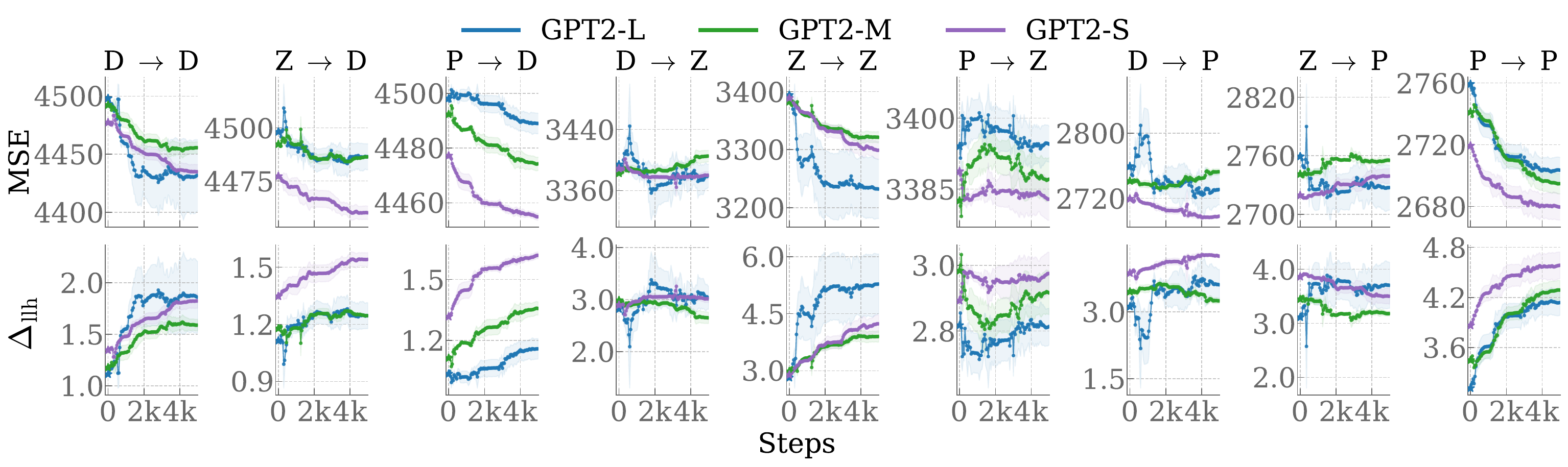}
  \caption {MSE and $\dll$ changes for \textbf{total reading durations}. Bands show standard errors across seeds.}
  \label{fig:mse_dll_trt_app}
\end{figure*}

\begin{figure*}[h!]
  \includegraphics[width=\linewidth]{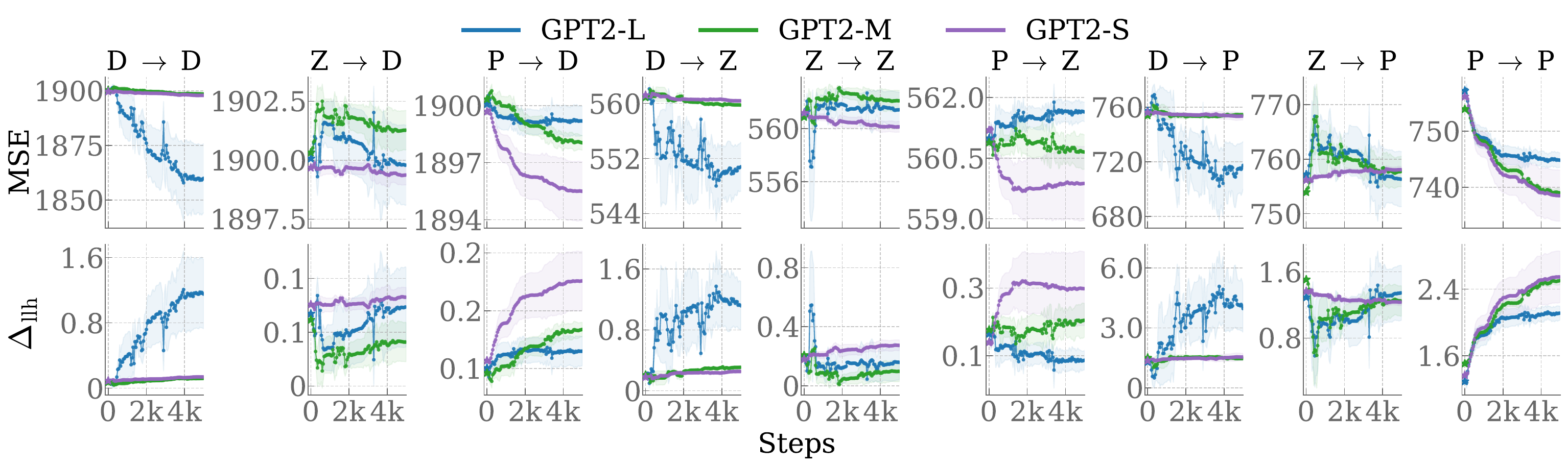}
  \caption {MSE and $\dll$ changes for \textbf{first fixation durations}. Bands show standard errors across seeds.}
  \label{fig:mse_dll_ffd_app}
\end{figure*}

Throughout this work, we have focused on predicting gaze durations. Here, we extend our analysis by fine-tuning models to predict total reading durations---which are the summed durations of all fixations on a unit $\unit$---and first fixation durations, which are the durations of the first fixation on a unit $\unit$. 
As shown in \Cref{tab:dll-table-trt-ffd-app}, models predicting total reading durations start at higher $\dll$ values, while models predicting first fixation durations start with lower $\dll$ values compared to those predicting gaze durations. On average, we observe lower percentage increases for total reading durations and lower increases for first fixation durations. Similar to the trajectories for gaze duration in \Cref{fig:mse-dll-change}, the trajectories for total reading durations in \Cref{fig:mse_dll_trt_app} show decreasing MSE and increasing $\dll$ for most configurations.
The trajectories for the first fixation durations (\Cref{fig:mse_dll_ffd_app}) are less consistent, particularly for models fine-tuned on the ZuCo corpus, where the MSE tends to increase throughout fine-tuning.

\begin{table}[h!] 
    \small
    \centering
    \begin{tabular}{l|ccc|ccc|ccc} 
    \toprule 
    \multicolumn{1}{c}{} & \multicolumn{3}{c}{GPT2-L} & \multicolumn{3}{c}{GPT2-M} & \multicolumn{3}{c}{GPT2-S} \\
    Data &  $\dll^{\text{start}}$ &  $\dll^{\text{max}}$ & \% Increase 
              &  $\dll^{\text{start}}$ &  $\dll^{\text{max}}$ & \% Increase 
              &  $\dll^{\text{start}}$ &  $\dll^{\text{max}}$ & \% Increase \\
    \midrule  
     \multicolumn{10}{c}{Total Reading Duration}\\
     \midrule
    D $\rightarrow$ D & $1.11$ & $1.93 \pm 0.33$ & 73.20 & $1.18$ & $1.62 \pm 0.07$ & 37.48 & $1.35$ & $1.83 \pm 0.07$ & 35.41 \\ 
    P $\rightarrow$ D & $1.11$ & $1.22 \pm 0.04$ & 9.20 & $1.18$ & $1.38 \pm 0.03$ & 17.26 & $1.35$ & $1.60 \pm 0.01$ & 18.62 \\ 
    Z $\rightarrow$ D & $1.11$ & $1.28 \pm 0.07$ & 15.01 & $1.18$ & $1.27 \pm 0.01$ & 7.83 & $1.35$ & $1.54 \pm 0.04$ & 14.47 \\ 
    D $\rightarrow$ P & $3.11$ & $3.79 \pm 0.20$ & 21.90 & $3.45$ & $3.63 \pm 0.06$ & 5.29 & $3.86$ & $4.27 \pm 0.05$ & 10.61 \\ 
    P $\rightarrow$ P & $3.11$ & $4.16 \pm 0.16$ & 33.64 & $3.45$ & $4.29 \pm 0.11$ & 24.43 & $3.86$ & $4.59 \pm 0.16$ & 18.70 \\ 
    Z $\rightarrow$ P & $3.11$ & $3.89 \pm 0.33$ & 24.95 & $3.45$ & $3.50 \pm 0.02$ & 1.42 & $3.86$ & $3.91 \pm 0.00$ & 1.33 \\ 
    D $\rightarrow$ Z & $2.82$ & $3.38 \pm 0.54$ & 19.93 & $2.98$ & $2.99 \pm 0.02$ & 0.19 & $2.89$ & $3.25 \pm 0.19$ & 12.47 \\ 
    P $\rightarrow$ Z & $2.82$ & $2.85 \pm 0.01$ & 1.31 & $2.98$ & $3.03 \pm 0.01$ & 1.64 & $2.89$ & $2.98 \pm 0.01$ & 3.10 \\ 
    Z $\rightarrow$ Z & $2.82$ & $5.28 \pm 0.79$ & 87.40 & $2.98$ & $3.92 \pm 0.14$ & 31.35 & $2.89$ & $4.23 \pm 0.23$ & 46.37 \\
     \midrule  
     \multicolumn{10}{c}{First Fixation Duration}\\
     \midrule
    D $\rightarrow$ D & $0.08$ & $1.21 \pm 0.47$ & 1397.26 & $0.07$ & $0.12 \pm 0.02$ & 64.10 & $0.09$ & $0.14 \pm 0.01$ & 52.02 \\ 
    P $\rightarrow$ D & $0.08$ & $0.11 \pm 0.02$ & 35.25 & $0.07$ & $0.13 \pm 0.01$ & 81.56 & $0.09$ & $0.20 \pm 0.04$ & 121.22 \\ 
    Z $\rightarrow$ D & $0.08$ & $0.10 \pm 0.02$ & 25.05 & $0.07$ & $0.07 \pm 0.00$ & 1.10 & $0.09$ & $0.10 \pm 0.01$ & 9.49 \\ 
    D $\rightarrow$ P & $1.29$ & $5.07 \pm 1.77$ & 292.36 & $1.51$ & $1.57 \pm 0.13$ & 3.87 & $1.37$ & $1.55 \pm 0.20$ & 13.46 \\ 
    P $\rightarrow$ P & $1.29$ & $2.13 \pm 0.08$ & 64.61 & $1.51$ & $2.50 \pm 0.05$ & 66.19 & $1.37$ & $2.55 \pm 0.31$ & 86.55 \\ 
    Z $\rightarrow$ P & $1.29$ & $1.44 \pm 0.41$ & 11.26 & $1.51$ & $1.51 \pm 0.01$ & 0.33 & $1.37$ & $1.38 \pm 0.02$ & 1.01 \\ 
    D $\rightarrow$ Z & $0.20$ & $1.37 \pm 0.46$ & 591.38 & $0.21$ & $0.31 \pm 0.01$ & 51.38 & $0.18$ & $0.25 \pm 0.02$ & 40.97 \\ 
    P $\rightarrow$ Z & $0.20$ & $0.21 \pm 0.01$ & 8.20 & $0.21$ & $0.23 \pm 0.01$ & 13.14 & $0.18$ & $0.32 \pm 0.06$ & 75.49 \\ 
    Z $\rightarrow$ Z & $0.20$ & $0.55 \pm 0.37$ & 175.48 & $0.21$ & $0.25 \pm 0.08$ & 21.36 & $0.18$ & $0.28 \pm 0.04$ & 52.88 \\ 
    \bottomrule
    \end{tabular}
    \caption{Mean start and maximum $\dll (10^{-2} \text{ nats})$ values using \textbf{total reading durations and first fixation durations}, including standard errors across random seeds, rounded to two decimal places.}
    \label{tab:dll-table-trt-ffd-app}
\end{table}

\subsection{Random Reading Times}
\label{asec:random-reading}
\begin{figure*}[h!]
  \includegraphics[width=\linewidth]{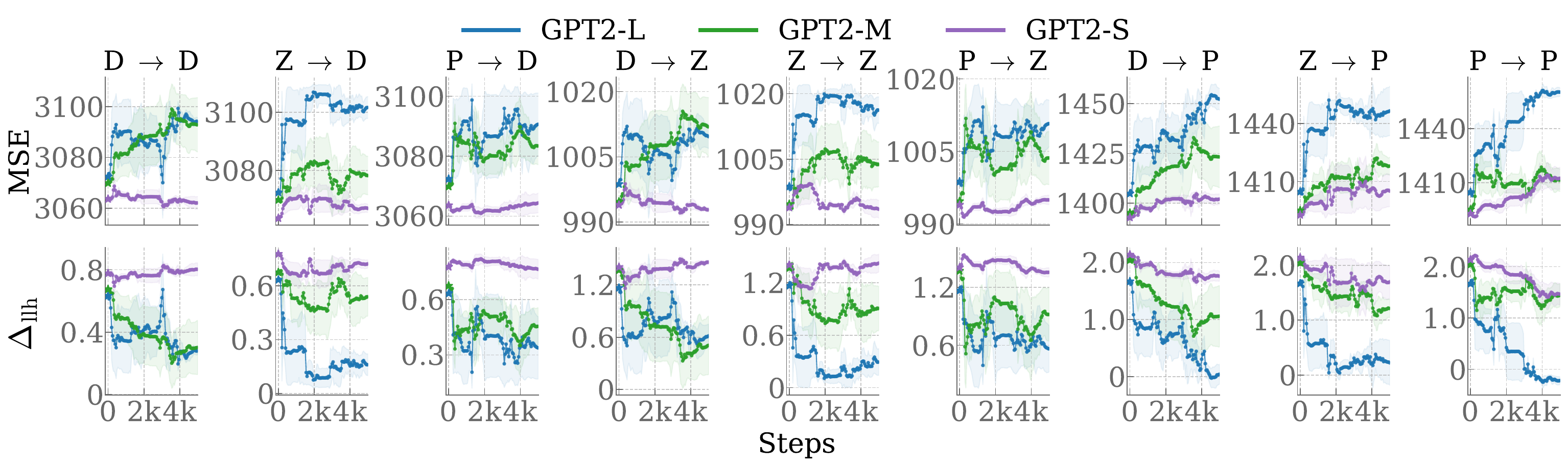}
  \caption {MSE and $\dll (10^{-2} \text{ nats})$ changes using \textbf{random reading times} sampled according to a normal distribution. Bands show the standard error across seeds. MSE increases, and $\dll$ decreases for most configurations, indicating that models become worse at predicting reading times throughout fine-tuning.}
  \label{fig:mse_dll_random_app}
\end{figure*}

Here we conduct additional experiments using random reading times to verify that the observed decreases in MSE and increases in $\dll$ are due to aligning language models with human reading times and not due to random noise or other confounding factors. Instead of fitting coefficients based on the reading times from the respective training dataset $\dataset$, we sample reading times from a Gaussian distribution, where the mean and standard deviation match those of the training dataset $\dataset$.
The results in \Cref{fig:mse_dll_random_app} show that fine-tuning on random reading times tends to have the opposite effect compared to fine-tuning on real reading times, leading to increasing MSE and decreasing $\dll$ throughout fine-tuning. Additionally, as in \Cref{tab:dll-table-app}, we compare the start and maximum $\dll$ values, as well as the percentage increases. \Cref{tab:dll-table-random-app} shows that fine-tuning with random reading times only leads to minimal or no increases. These results show that random reading times do not improve language models at predicting reading times and confirm the effectiveness of our technique for aligning models to human reading times.
\begin{table}[ht!] 
    \small
    \centering
    \begin{tabular}{l|ccc|ccc|ccc} 
    \toprule 
    \multicolumn{1}{c}{} & \multicolumn{3}{c}{GPT2-L} & \multicolumn{3}{c}{GPT2-M} & \multicolumn{3}{c}{GPT2-S} \\
    Data &  $\dll^{\text{start}}$ &  $\dll^{\text{max}}$ & \% Increase 
              &  $\dll^{\text{start}}$ &  $\dll^{\text{max}}$ & \% Increase 
              &  $\dll^{\text{start}}$ &  $\dll^{\text{max}}$ & \% Increase \\ 
     \midrule  
    D $\rightarrow$ D & $0.63$ & $0.67 \pm 0.23$ & 6.31 & $0.68$ & $0.68 \pm 0.00$ & 0.00 & $0.78$ & $0.83 \pm 0.03$ & 6.42 \\ 
    P $\rightarrow$ D & $0.63$ & $0.66 \pm 0.00$ & 3.84 & $0.68$ & $0.68 \pm 0.00$ & 0.00 & $0.78$ & $0.82 \pm 0.03$ & 5.67 \\ 
    Z $\rightarrow$ D & $0.63$ & $0.64 \pm 0.01$ & 0.88 & $0.68$ & $0.68 \pm 0.01$ & 1.30 & $0.78$ & $0.78 \pm 0.00$ & 0.00 \\ 
    D $\rightarrow$ P & $1.67$ & $1.67 \pm 0.00$ & 0.00 & $2.04$ & $2.06 \pm 0.01$ & 0.56 & $2.14$ & $2.14 \pm 0.00$ & 0.00 \\ 
    P $\rightarrow$ P & $1.67$ & $1.74 \pm 0.01$ & 4.06 & $2.04$ & $2.04 \pm 0.00$ & 0.00 & $2.14$ & $2.22 \pm 0.02$ & 3.87 \\ 
    Z $\rightarrow$ P & $1.67$ & $1.71 \pm 0.02$ & 2.44 & $2.04$ & $2.08 \pm 0.01$ & 1.66 & $2.14$ & $2.14 \pm 0.01$ & 0.03 \\ 
    D $\rightarrow$ Z & $1.17$ & $1.17 \pm 0.00$ & 0.00 & $1.37$ & $1.37 \pm 0.00$ & 0.17 & $1.41$ & $1.50 \pm 0.04$ & 6.48 \\ 
    P $\rightarrow$ Z & $1.17$ & $1.20 \pm 0.02$ & 3.20 & $1.37$ & $1.37 \pm 0.00$ & 0.00 & $1.41$ & $1.53 \pm 0.01$ & 9.14 \\ 
    Z $\rightarrow$ Z & $1.17$ & $1.20 \pm 0.01$ & 2.98 & $1.37$ & $1.42 \pm 0.03$ & 3.40 & $1.41$ & $1.45 \pm 0.04$ & 2.86 \\  
    \bottomrule
    \end{tabular}
    \caption{Mean start and maximum $\dll (10^{-2} \text{ nats})$ values for \textbf{random reading times}, including standard errors across random seeds, rounded to two decimal places.}
    \label{tab:dll-table-random-app}
\end{table}

\clearpage

\section{Coefficients}
\label{asec:all_coeffs}
\begin{figure*}[h!]
\begin{subfigure}[b]{0.49\linewidth}
  \includegraphics[width=\linewidth]{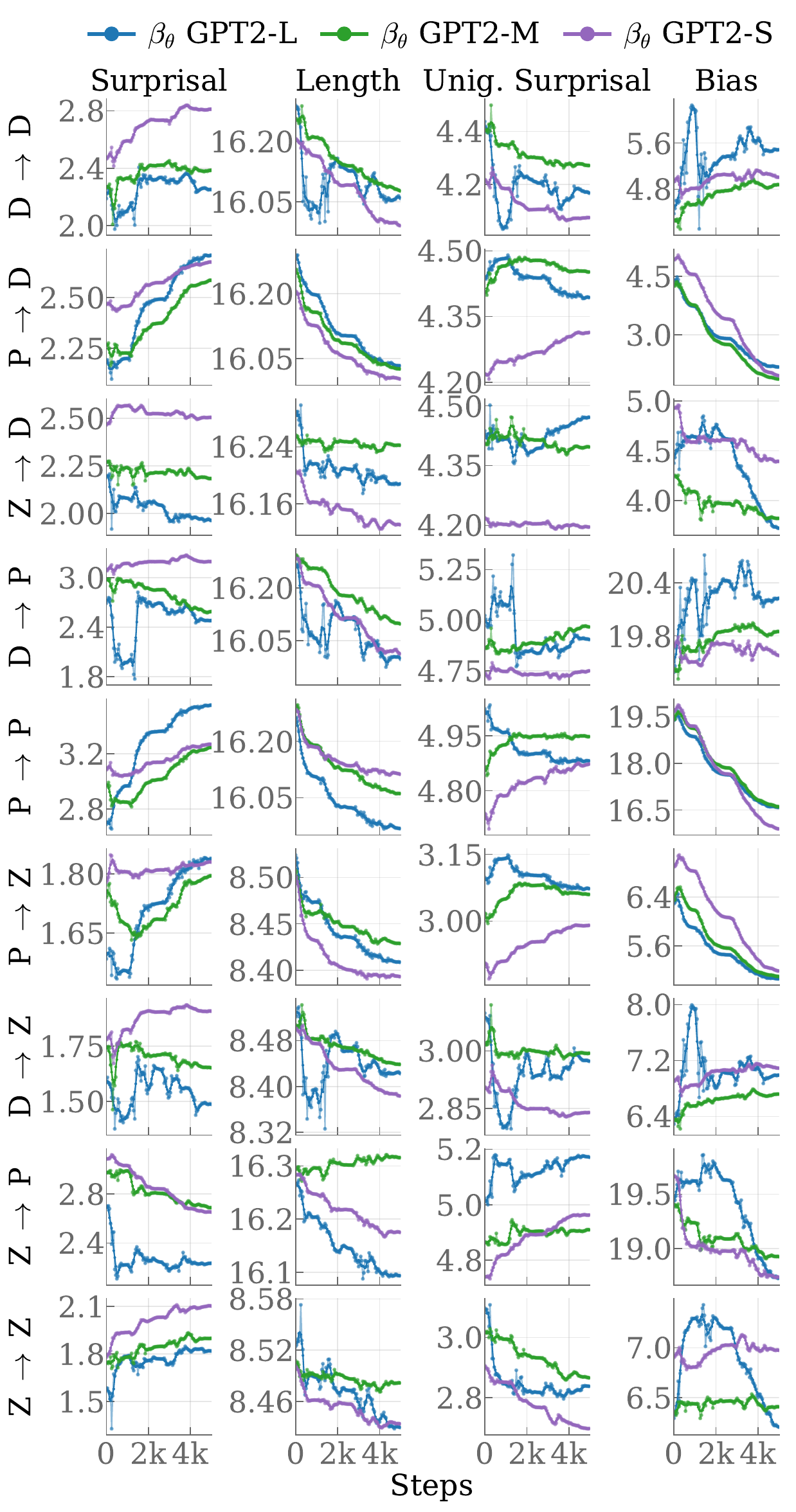}
  \caption{Unregularized objective ($\klReg = 0$)}
  \label{fig:app_0_coefficients}
  \end{subfigure}
  \hspace{0.01\linewidth}
  \begin{subfigure}[b]{0.49\linewidth}
  \includegraphics[width=\linewidth]{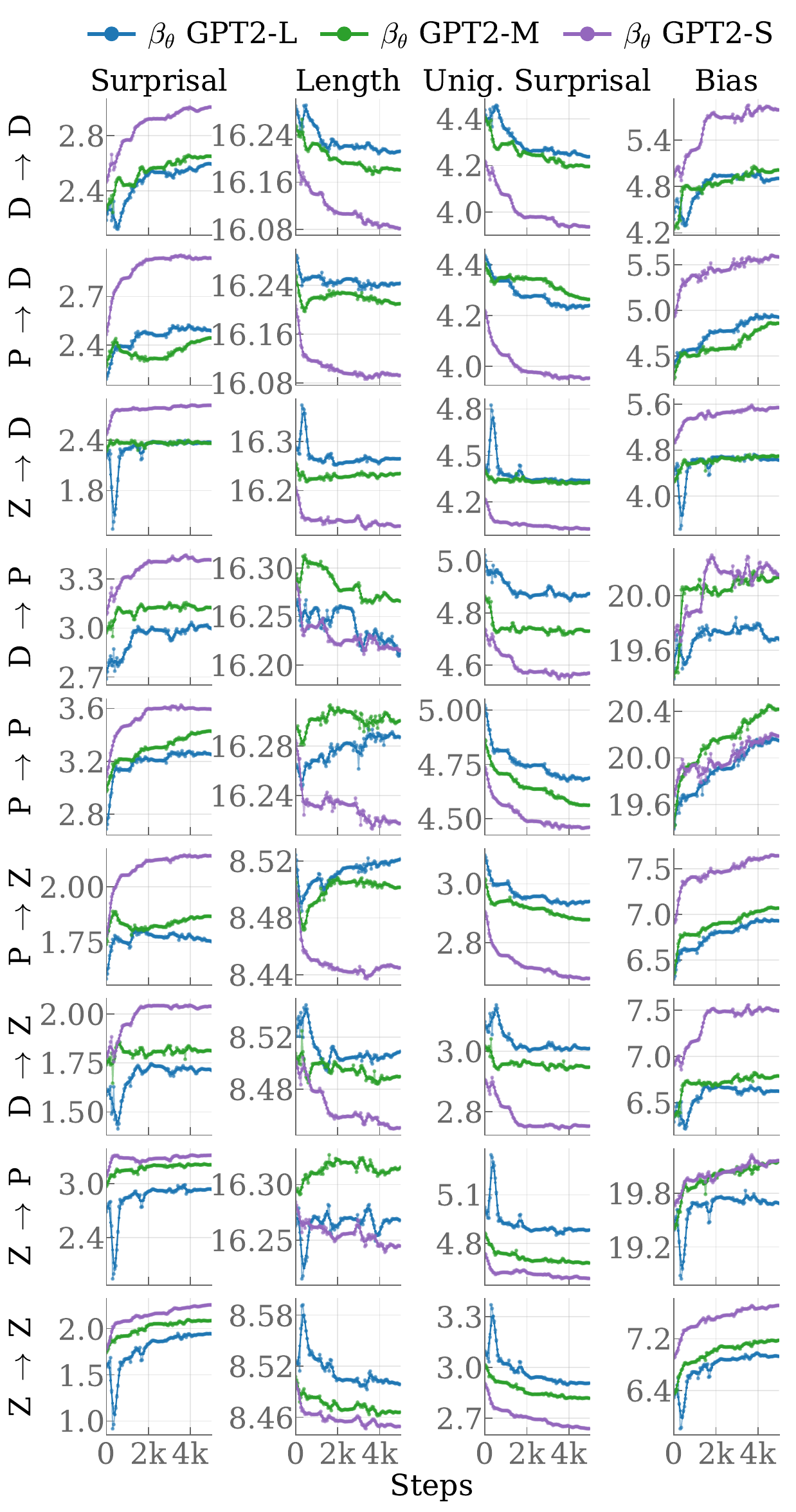}
  \caption{Regularized objective ($\klReg = 500$)}
  \label{fig:app_500_coefficients}
  \end{subfigure}
  \caption {\textbf{Mean coefficients of unit-level features over fine-tuning for all data splits.} Smoothed values (window size 5) are shown, with unsmoothed values in a pale version of the color.}
  \label{fig:app_coefficients}
\end{figure*}

\Cref{fig:app_0_coefficients} shows the trajectory of coefficients for all data splits during fine-tuning using the unregularized objective. While the results for unigram surprisal and bias coefficients are mixed, the coefficients for surprisal tend to increase during fine-tuning, and those for length tend to decrease. These trends are more consistent in experiments with the largest increases in $\dll$, i.e.,  where the training and test splits come from the same datasets. In contrast, the coefficients in experiments involving domain shift (where the training and test splits come from different datasets) do not have such a clear pattern, particularly those without consistent increases in $\dll$ (Dundee $\rightarrow$ Provo and ZuCo $\rightarrow$ Provo).

Additionally, we visualize all coefficients from models trained with the regularized objective ($\klReg = 500$). As shown in \Cref{fig:mse_dll_500_app}, regularization leads to lower $\dll$ increases but more consistent trajectories across configurations, which is why we also hypothesize more consistent trajectories of the coefficients. 
As shown in \Cref{fig:app_500_coefficients}, the coefficients for surprisal and bias show a clear upward trend throughout fine-tuning, while the coefficients for unigram surprisal show a clear downward trend.

\clearpage

\section{Text-Generation}
\label{asec:text-generation}
Here we expand on the text-generation experiments described in \Cref{ssec:text-gen}. We sample 500 prefixes $\prompt$, each consisting of the first three words from the CNN/DailyMail dataset \citep{hermann-etal-2015-teaching, see-etal-2017-get}, and generate completions $\sampleString \sim \model(\cdot \mid \prompt)$ of up to 50 tokens using our fine-tuned language models. For a completion $\sampleString$ of length $N$, we write $\unit_n$ to denote the $n^{\text{th}}$ unit of $\sampleString$. Further, let $\context_n$ be the context of $\unit_n$ in $\sampleString$, including the prefix $\prompt$. We exclude short completions with $\vert\sampleString\vert < 3$ and estimate surprisal with a separate language model, Pythia-70m \citep{biderman2023pythia} using the code from \citet{pimentel2024computeprobabilityword}.\footnote{\url{https://github.com/tpimentelms/probability-of-a-word}} Recently, \citet{oh2024leadingwhitespaceslanguagemodels, pimentel2024computeprobabilityword} have argued that leading whitespaces from tokenization pose a confound to surprisal calculations and that the probability of trailing whitespaces should be included instead. However, we do not include a unit's trailing whitespace in our surprisal calculation; see \citet{giulianelli2024propertreatmenttokenizationpsycholinguistics}.
Then, following previous work \citep{meister-etal-2021-revisiting, clark-etal-2023-cross}, we measure the uniformity of information of a generated completion $\sampleString$ given the prefix $\prompt$ using the mean surprisal variance
\begin{equation}
  \label{eqn:surprisal-variance}
    \surprisalvariance(\sampleString \mid \prompt)  = \frac{1}{N}\sum_{n=1}^{N}(\surprisal_{\btheta}(\unit_n \mid \context_n) - \mu_{\surprisal_{\btheta}}(\sampleString \mid \prompt))^2,
\end{equation}
where the mean surprisal $\mu_{\surprisal_{\btheta}}(\sampleString \mid \prompt)$ is given by
\begin{equation}
  \label{eqn:mean_surprisal}
    \mu_{\surprisal_{\btheta}}(\sampleString \mid \prompt)  = \frac{1}{N}\sum_{n=1}^{N}\surprisal_{\btheta}(\unit_n \mid \context_n).
\end{equation}
Additionally, we calculate the mean local surprisal variance:
\begin{equation}
  \label{eqn:local-surprisal-variance}
    \localsurprisalvariance(\sampleString \mid \prompt)  = \frac{1}{N-1}\sum_{n=2}^{N}\big(\surprisal_{\btheta}(\unit_n \mid \context_n) - \surprisal_{\btheta}(\unit_{n-1} \mid \context_{{n-1}})\big)^2.
\end{equation}
To evaluate the diversity of the generations, we calculate the mean unique $n$-gram ratio ($n$-Gram\%) over completions $\sampleString$. 
In \Cref{tab:generation_metrics_app}, we report the mean surprisal, surprisal variance, local surprisal variance, and unique $n$-gram ratios across all data splits. For models fine-tuned without regularization ($\klReg = 0$), surprisal variance and local variance tend to increase compared to the pretrained models, with a few exceptions, particularly for GPT2-M, where the variance and local variance remain close to the pretrained models. Overall, this indicates that information becomes less uniformly distributed. However, under KL regularization ($\klReg = 500$), this trend is reversed, and we observe more uniform information in the generated text with the exception being GPT2-L: Dundee $\rightarrow$ Dundee. 
Additionally, we observe a decline in the ratio of unique unigrams compared to the pretrained models, indicating that fine-tuned models generate more repetitive text. However, diversity and uniformity of information are not necessarily linked, as models fine-tuned without regularization tend to generate less diverse and less uniform text. 

\label{asec:text-gen}
\begin{table}[h!]
\small
\centering
\begin{tabular}{lccc|ccc}
\toprule
Model & $\downarrow \mu_{\surprisal_{\btheta}}$ & $\downarrow \surprisalvariance$ & $\downarrow \localsurprisalvariance$ & $\uparrow$1-Gram\% & $\uparrow$2-Gram\% & $\uparrow$3-Gram\% \\
\midrule
GPT2-L & $3.00$ & $6.69$ & $14.25$ & $84.84$ & $94.16$ & $95.86$ \\
GPT2-M & $2.94$ & $7.67$ & $16.62$ & $84.39$ & $93.07$ & $94.83$ \\
GPT2-S & $2.62$ & $5.70$ & $11.65$ & $82.53$ & $91.63$ & $93.35$ \\
\midrule
\multicolumn{7}{c}{\textbf{$\klReg = 0$}} \\ 
\midrule
GPT2-L D $\rightarrow$ D & $3.38_{0.35}$ & $10.17_{2.93}$ & $20.58_{6.94}$ & $67.07_{14.68}$ & $74.79_{16.37}$ & $76.71_{16.32}$ \\
GPT2-L P $\rightarrow$ D & $2.96_{0.03}$ & $11.29_{1.74}$ & $23.76_{4.66}$ & $68.88_{1.50}$ & $79.78_{1.48}$ & $82.79_{1.53}$ \\
GPT2-L Z $\rightarrow$ D & $3.49_{0.21}$ & $9.71_{1.35}$ & $20.15_{3.03}$ & $78.02_{4.05}$ & $92.40_{2.04}$ & $95.48_{1.26}$ \\
GPT2-L D $\rightarrow$ P & $3.03_{0.45}$ & $9.40_{2.49}$ & $19.26_{4.80}$ & $76.41_{3.92}$ & $85.80_{4.50}$ & $88.17_{4.39}$ \\
GPT2-L P $\rightarrow$ P & $2.97_{0.04}$ & $11.27_{1.74}$ & $23.81_{4.64}$ & $68.91_{1.76}$ & $79.91_{1.66}$ & $82.88_{1.65}$ \\
GPT2-L Z $\rightarrow$ P & $3.71_{0.27}$ & $11.22_{2.87}$ & $23.38_{5.82}$ & $86.74_{0.79}$ & $96.36_{0.42}$ & $97.98_{0.49}$ \\
GPT2-L D $\rightarrow$ Z & $3.24_{0.20}$ & $7.72_{1.57}$ & $14.48_{4.86}$ & $61.65_{19.46}$ & $68.38_{21.10}$ & $70.16_{21.20}$ \\
GPT2-L P $\rightarrow$ Z & $2.97_{0.03}$ & $6.60_{0.08}$ & $14.06_{0.19}$ & $82.93_{1.91}$ & $92.89_{1.27}$ & $95.00_{0.86}$ \\
GPT2-L Z $\rightarrow$ Z & $3.99_{0.12}$ & $13.85_{0.98}$ & $27.72_{2.46}$ & $84.60_{1.91}$ & $94.71_{2.05}$ & $96.39_{1.68}$ \\
\midrule
GPT2-M D $\rightarrow$ D & $3.27_{0.03}$ & $8.94_{0.25}$ & $18.22_{0.71}$ & $80.82_{0.35}$ & $91.37_{0.63}$ & $93.82_{0.73}$ \\
GPT2-M P $\rightarrow$ D & $3.04_{0.07}$ & $7.81_{0.47}$ & $15.85_{1.21}$ & $75.82_{0.24}$ & $87.77_{0.17}$ & $90.91_{0.26}$ \\
GPT2-M Z $\rightarrow$ D & $3.27_{0.24}$ & $7.95_{0.92}$ & $16.70_{1.74}$ & $83.37_{1.59}$ & $93.87_{1.44}$ & $95.81_{1.15}$ \\
GPT2-M D $\rightarrow$ P & $2.83_{0.09}$ & $6.96_{1.02}$ & $14.73_{2.40}$ & $81.05_{1.78}$ & $90.15_{1.62}$ & $92.25_{1.45}$ \\
GPT2-M P $\rightarrow$ P & $2.99_{0.04}$ & $7.81_{0.46}$ & $14.52_{1.42}$ & $67.51_{0.78}$ & $78.71_{0.55}$ & $82.19_{0.68}$ \\
GPT2-M Z $\rightarrow$ P & $3.08_{0.09}$ & $7.25_{0.60}$ & $15.18_{1.32}$ & $84.68_{0.70}$ & $93.70_{0.89}$ & $95.37_{0.81}$ \\
GPT2-M D $\rightarrow$ Z & $2.89_{0.02}$ & $7.61_{0.80}$ & $15.62_{1.20}$ & $81.18_{0.69}$ & $91.21_{0.98}$ & $93.46_{1.00}$ \\
GPT2-M P $\rightarrow$ Z & $2.92_{0.03}$ & $7.45_{0.45}$ & $15.53_{0.79}$ & $76.63_{4.35}$ & $87.13_{3.91}$ & $89.82_{3.52}$ \\
GPT2-M Z $\rightarrow$ Z & $3.78_{0.13}$ & $12.12_{1.42}$ & $24.12_{2.59}$ & $84.95_{0.86}$ & $94.30_{0.85}$ & $96.00_{0.80}$ \\
\midrule
GPT2-S D $\rightarrow$ D & $2.72_{0.13}$ & $6.27_{0.52}$ & $13.09_{1.52}$ & $81.14_{0.95}$ & $89.13_{1.19}$ & $91.02_{1.22}$ \\
GPT2-S P $\rightarrow$ D & $2.94_{0.05}$ & $7.03_{0.31}$ & $13.92_{0.58}$ & $71.91_{2.60}$ & $83.59_{2.19}$ & $86.61_{1.83}$ \\
GPT2-S Z $\rightarrow$ D & $2.89_{0.13}$ & $7.24_{0.79}$ & $14.02_{1.19}$ & $79.21_{2.10}$ & $89.31_{1.04}$ & $91.61_{0.73}$ \\
GPT2-S D $\rightarrow$ P & $2.72_{0.10}$ & $6.54_{0.27}$ & $13.05_{0.59}$ & $77.08_{3.04}$ & $85.98_{2.90}$ & $88.30_{2.66}$ \\
GPT2-S P $\rightarrow$ P & $2.95_{0.07}$ & $6.91_{0.25}$ & $13.79_{0.93}$ & $72.17_{1.99}$ & $83.91_{1.42}$ & $86.87_{1.19}$ \\
GPT2-S Z $\rightarrow$ P & $2.62_{0.04}$ & $5.49_{0.32}$ & $11.21_{0.49}$ & $81.92_{1.03}$ & $90.86_{0.94}$ & $92.63_{0.76}$ \\
GPT2-S D $\rightarrow$ Z & $2.54_{0.07}$ & $5.81_{0.25}$ & $11.25_{0.30}$ & $76.73_{2.73}$ & $85.56_{2.60}$ & $87.84_{2.33}$ \\
GPT2-S P $\rightarrow$ Z & $2.58_{0.01}$ & $5.23_{0.27}$ & $10.59_{0.73}$ & $77.17_{2.97}$ & $87.90_{2.13}$ & $90.46_{1.60}$ \\
GPT2-S Z $\rightarrow$ Z & $2.96_{0.13}$ & $7.79_{1.02}$ & $15.40_{2.02}$ & $79.61_{1.94}$ & $89.08_{2.34}$ & $91.21_{2.32}$ \\
\midrule
\multicolumn{7}{c}{\textbf{$\klReg = 500$}} \\ 
\midrule
GPT2-L D $\rightarrow$ D & $3.01_{0.04}$ & $8.05_{0.77}$ & $16.54_{1.34}$ & $82.38_{0.18}$ & $92.39_{0.40}$ & $94.26_{0.43}$ \\
GPT2-L P $\rightarrow$ D & $2.48_{0.03}$ & $5.23_{0.40}$ & $10.35_{0.65}$ & $76.24_{0.66}$ & $86.37_{0.62}$ & $88.85_{0.60}$ \\
GPT2-L Z $\rightarrow$ D & $2.70_{0.02}$ & $5.64_{0.01}$ & $12.00_{0.10}$ & $80.91_{0.63}$ & $91.11_{0.39}$ & $93.15_{0.33}$ \\
GPT2-L D $\rightarrow$ P & $2.85_{0.01}$ & $5.97_{0.07}$ & $12.62_{0.16}$ & $83.48_{0.60}$ & $92.49_{0.63}$ & $94.23_{0.63}$ \\
GPT2-L P $\rightarrow$ P & $2.46_{0.03}$ & $5.14_{0.35}$ & $9.95_{0.55}$ & $76.75_{0.79}$ & $86.68_{0.50}$ & $89.01_{0.38}$ \\
GPT2-L Z $\rightarrow$ P & $2.82_{0.05}$ & $6.12_{0.31}$ & $13.06_{0.72}$ & $83.11_{0.83}$ & $92.96_{0.61}$ & $94.75_{0.52}$ \\
GPT2-L D $\rightarrow$ Z & $2.88_{0.03}$ & $6.37_{0.19}$ & $13.64_{0.41}$ & $83.04_{0.71}$ & $91.95_{0.71}$ & $93.69_{0.63}$ \\
GPT2-L P $\rightarrow$ Z & $2.34_{0.01}$ & $4.46_{0.07}$ & $8.84_{0.09}$ & $73.86_{0.99}$ & $84.26_{0.90}$ & $86.96_{0.84}$ \\
GPT2-L Z $\rightarrow$ Z & $2.75_{0.02}$ & $5.61_{0.09}$ & $11.96_{0.30}$ & $81.12_{0.20}$ & $91.51_{0.42}$ & $93.63_{0.48}$ \\
\midrule
GPT2-M D $\rightarrow$ D & $2.85_{0.02}$ & $6.63_{0.35}$ & $13.98_{0.78}$ & $83.22_{1.08}$ & $92.17_{0.33}$ & $94.09_{0.14}$ \\
GPT2-M P $\rightarrow$ D & $2.70_{0.03}$ & $5.73_{0.54}$ & $11.74_{1.14}$ & $79.41_{0.88}$ & $90.14_{0.48}$ & $92.64_{0.31}$ \\
GPT2-M Z $\rightarrow$ D & $2.72_{0.04}$ & $6.07_{0.17}$ & $12.76_{0.30}$ & $81.52_{0.78}$ & $91.55_{0.64}$ & $93.72_{0.56}$ \\
GPT2-M D $\rightarrow$ P & $2.80_{0.07}$ & $6.79_{0.83}$ & $14.41_{2.03}$ & $83.00_{1.39}$ & $91.46_{1.04}$ & $93.41_{0.91}$ \\
GPT2-M P $\rightarrow$ P & $2.46_{0.03}$ & $4.54_{0.17}$ & $9.16_{0.42}$ & $75.91_{0.85}$ & $86.71_{0.59}$ & $89.65_{0.56}$ \\
GPT2-M Z $\rightarrow$ P & $2.72_{0.03}$ & $6.00_{0.04}$ & $12.59_{0.10}$ & $81.54_{1.14}$ & $91.27_{0.78}$ & $93.28_{0.76}$ \\
GPT2-M D $\rightarrow$ Z & $2.80_{0.06}$ & $6.75_{0.14}$ & $14.05_{0.34}$ & $81.90_{1.04}$ & $91.21_{0.82}$ & $93.18_{0.80}$ \\
GPT2-M P $\rightarrow$ Z & $2.57_{0.09}$ & $5.32_{0.46}$ & $10.82_{0.98}$ & $77.08_{1.73}$ & $88.04_{1.19}$ & $90.78_{0.98}$ \\
GPT2-M Z $\rightarrow$ Z & $2.72_{0.03}$ & $5.88_{0.05}$ & $12.17_{0.21}$ & $81.94_{0.91}$ & $91.53_{0.55}$ & $93.52_{0.51}$ \\
\midrule
GPT2-S D $\rightarrow$ D & $2.37_{0.09}$ & $4.50_{0.11}$ & $9.30_{0.17}$ & $76.56_{2.02}$ & $85.95_{2.28}$ & $87.90_{2.23}$ \\
GPT2-S P $\rightarrow$ D & $2.13_{0.08}$ & $4.30_{0.28}$ & $8.62_{0.70}$ & $68.97_{2.33}$ & $78.38_{2.49}$ & $81.41_{2.41}$ \\
GPT2-S Z $\rightarrow$ D & $2.34_{0.02}$ & $4.37_{0.05}$ & $8.84_{0.13}$ & $74.31_{1.18}$ & $85.19_{0.79}$ & $87.66_{0.66}$ \\
GPT2-S D $\rightarrow$ P & $2.40_{0.08}$ & $4.60_{0.23}$ & $9.69_{0.43}$ & $79.17_{2.12}$ & $87.65_{1.83}$ & $89.28_{1.69}$ \\
GPT2-S P $\rightarrow$ P & $2.15_{0.04}$ & $4.11_{0.10}$ & $8.34_{0.19}$ & $69.54_{0.43}$ & $79.13_{0.51}$ & $82.08_{0.50}$ \\
GPT2-S Z $\rightarrow$ P & $2.45_{0.05}$ & $5.07_{0.36}$ & $10.47_{0.57}$ & $79.42_{0.94}$ & $88.60_{0.93}$ & $90.52_{0.79}$ \\
GPT2-S D $\rightarrow$ Z & $2.36_{0.05}$ & $4.52_{0.12}$ & $9.36_{0.29}$ & $77.73_{1.12}$ & $86.73_{1.04}$ & $88.58_{1.02}$ \\
GPT2-S P $\rightarrow$ Z & $2.17_{0.11}$ & $4.22_{0.20}$ & $8.65_{0.46}$ & $70.47_{2.41}$ & $79.94_{2.76}$ & $82.95_{2.69}$ \\
GPT2-S Z $\rightarrow$ Z & $2.35_{0.03}$ & $4.97_{0.18}$ & $10.25_{0.47}$ & $75.92_{0.68}$ & $85.66_{0.77}$ & $87.76_{0.88}$ \\
\bottomrule
\end{tabular}
\caption{Full evaluation results for completions generated on prefixes sampled from CNN/Dailymail.}
\label{tab:generation_metrics_app}

\end{table}

\end{document}